%% file: main.tex
\documentclass[10pt,twocolumn,letterpaper]{article}

\usepackage[pagenumbers]{cvpr}

\input{preamble.tex}

\definecolor{linkblue}{rgb}{0.21,0.49,0.74}
\usepackage[breaklinks,colorlinks,allcolors=linkblue]{hyperref}

\title{NNS-GS: Non-Local Neural Scattering for Relightable Gaussian Splatting}

\input{authors.tex}
\date{}
\hypersetup{pdftitle={NNS-GS: Non-Local Neural Scattering for Relightable Gaussian Splatting},pdfauthor={Shiyu Liang, Yang Gao, Lili Wang, Aimin Hao, Beibei Wang},pdfsubject={Research preprint}}

\begin{document}
\maketitle
\input{sec_0_abstract.tex}
\input{sec_1_intro.tex}

\input{sec_2_related_work.tex}
\input{sec_3_method.tex}

\input{sec_4_experiments.tex}

\input{sec_5_conclusion.tex}

{
    \small
    \bibliographystyle{ieeenat_fullname}
    \bibliography{intro_refs,related_refs,expanded_refs}
}

\end{document}


\maketitle
\appendix
\renewcommand{\thetable}{S\arabic{table}}
\renewcommand{\thefigure}{S\arabic{figure}}
\renewcommand{\theequation}{S\arabic{equation}}
\raggedbottom

We present NNS-GS, a framework for reconstructing and relighting translucent objects through non-local neural scattering and geometry-guided reflection. This supplementary material provides further analysis of the framework. Section~\ref{sec:supp_implementation} details the implementation choices. Section~\ref{sec:supp_validation} provides additional experimental validation. Section~\ref{sec:limitations} discusses remaining challenges and future directions.

\input{supplementary_sec_implementation.tex}

\input{supplementary_sec_additional_results.tex}

\input{supplementary_sec_relighting.tex}
\input{supplementary_sec_editing.tex}
\input{supplementary_sec_limitations.tex}

\FloatBarrier

{
\small
\bibliographystyle{ieeenat_fullname}
\bibliography{supplementary_references}
}

%% file: preamble.tex
\usepackage{cuted}
\usepackage{colortbl}

\usepackage{currfile}

\usepackage{caption}

%% file: authors.tex
\author{Shiyu Liang\textsuperscript{1}\quad Yang Gao\textsuperscript{1}\quad Lili Wang\textsuperscript{1}\quad Aimin Hao\textsuperscript{1}\quad Beibei Wang\textsuperscript{2}\\[4pt]
\textsuperscript{1}Beihang University\qquad \textsuperscript{2}Nanjing University}

%% file: sec_0_abstract.tex
\begin{abstract}
Creating relightable 3D assets of translucent objects is challenging because changing the illumination at one surface location can affect the appearance elsewhere through subsurface scattering. Existing Gaussian-based scattering models typically predict scattering from local features, making this dependence difficult to capture. We present Non-Local Neural Scattering for Gaussian Splatting (NNS-GS), which learns non-local scattering together with geometry-guided local reflection. The scattering module learns how illumination from light-visible surface regions in light space contributes to the shading point in view space, conditioned on regional diffuse responses and spatial relationships. The reflection module supplies these diffuse responses and uses additional geometric guidance to better capture specular highlights and sharp shadow boundaries. To reduce component mixing and support controllable scattering appearance, anchor-based progressive training establishes the diffuse–scattering base before introducing specular reflection and supervises the components separately. Under calibrated point lighting, NNS-GS improves mean PSNR over existing methods by 2.47 dB and 0.81 dB on the synthetic and real translucent scenes, respectively. Additional synthetic comparisons further show that NNS-GS better matches the reference when adjusting scattering attenuation.
\end{abstract}

%% file: sec_1_intro.tex
\section{Introduction}
\label{sec:intro}
Creating relightable 3D translucent objects is important for realistic rendering in games, virtual reality, and product visualization. This task is challenging because subsurface scattering (SSS) allows light entering one surface region to scatter within the object and emerge elsewhere~\cite{jensen2001subsurface}. Predicting appearance under new lighting, therefore, requires accounting for illumination across the surface and its transport through the object, beyond the local lighting conditions at each point.

Existing relighting methods based on neural scene representations, including neural radiance fields (NeRFs)~\cite{mildenhall2020nerf} and 3D Gaussian splatting (3DGS)~\cite{kerbl2023gaussians}, separate geometry, materials, and illumination to enable physically based relighting~\cite{srinivasan2021nerv,zhang2021nerfactor,jin2023tensoir,gao2024r3dg,liang2024gsir,bi2024gs3,gu2025irgs} or learn lighting-dependent appearance to reproduce complex material responses~\cite{gao2020dnl,zeng2023nrhints,kuang2024olat,fan2025rng}. These methods do not explicitly model SSS and can miss the soft illumination of translucent objects. SSS-GS~\cite{dihlmann2024sssgs} and SSD-GS~\cite{zheng2026ssdgs} model SSS within Gaussian representations using a learned scattering residual and a dipole-based formulation, respectively. However, their scattering modules evaluate per-Gaussian inputs without explicitly aggregating information from other surface locations. This limits their ability to accurately reproduce transmission effects under new lighting.

In this paper, we present Non-Local Neural Scattering for Gaussian Splatting (NNS-GS), a novel framework built on a non-local neural scattering representation that couples surface illumination with relative geometry.
To capture non-local scattering, we use the diffuse responses in light space as illumination cues at other surface locations with their 3D distances, allowing the model to learn how illumination across the surface contributes to local scattered appearance in view space.
Alongside scattering, local surface reflection introduces sharp highlights and shadow boundaries, whose rapid angular and spatial variations can be difficult to reproduce. Therefore, our proposed geometry-guided neural reflection module introduces additional geometric cues to guide the prediction of these details: learned local frames provide directional context for highlight modeling, while light-space depth differences provide occlusion-related cues for refining shadow boundaries.

During training, the flexible specular branch can absorb appearance already explained by diffuse reflection and scattering. Such mixing can compromise editing even when reconstruction quality remains high. We therefore introduce anchor-based progressive training, which anchors the diffuse–scattering base to its pre-specular estimate and supervises the specular branch with the positive residual relative to that anchor. On the translucent-material dataset, NNS-GS improves mean PSNR over existing methods by 2.47 dB on synthetic scenes and 0.81 dB on real scenes, supporting controlled scattering appearance adjustment and rendering performance at about 60 FPS.

Overall, our main contributions are:
\begin{itemize}
    \item a novel Gaussian framework (NNS-GS) that achieves more faithful relighting of translucent objects and supports plausible adjustments to scattering appearance,
    \item a dual-space non-local neural scattering model that learns how illumination at multiple surface locations jointly contributes to local scattered appearance, and
    \item a geometry-guided neural reflection model that uses additional geometric cues to enhance fine highlight detail and refine cast shadow boundaries.
\end{itemize}

%% file: sec_2_related_work.tex
\section{Related Work}
\label{sec:related_work}

\subsection{3D Representations and Relighting}
NeRF~\cite{mildenhall2020nerf} represents scenes as continuous fields, while 3DGS~\cite{kerbl2023gaussians} uses explicit Gaussian primitives for efficient rendering.
Both support high-quality reconstruction and novel-view synthesis, but their original formulations capture appearance under observed lighting rather than separate material and illumination.
Relighting methods based on NeRF and related neural fields address this limitation by separating reflectance from illumination~\cite{boss2021nerd,zhang2021nerfactor,zhang2021physg}, modeling visibility and indirect light~\cite{srinivasan2021nerv,jin2023tensoir}, or learning how appearance varies with lighting from captured images~\cite{zeng2023nrhints}. Recent work uses physical constraints to better separate material properties from lighting~\cite{wu2025pbrnerf} and time-resolved measurements to distinguish direct from indirect light~\cite{malik2025propagating}.

With 3DGS, relighting approaches extend explicit scene representations with material properties and shading models for relighting~\cite{jiang2024gaussianshader,liang2024gsir,gao2024r3dg}.
Subsequent advances improve geometry and reflectance estimation~\cite{sun2025svgir,zhu2025sdfgs}, account for visibility and indirect illumination~\cite{chen2025gigs,gu2025irgs}, and reduce material--lighting ambiguity through cross-environment observations and stronger lighting or material priors~\cite{li2025recap,du2025gsid,zhang2026irhgp,langsteiner2026matspray}.
For calibrated point-light capture, GS$^3$ and related methods provide flexible directional appearance models~\cite{bi2024gs3,kuang2024olat}, while RNG~\cite{fan2025rng} learns neural reflectance from latent Gaussian features.
Together, these methods establish effective reconstruction and relighting frameworks, predominantly for surface reflection from opaque materials.

\subsection{Subsurface Scattering}
SSS contributes to the soft appearance of translucent materials. It connects the appearance of a surface point to light entering the object at other locations.
The bidirectional surface scattering reflectance distribution function (BSSRDF) describes this spatial relationship, while diffusion models enable efficient evaluation~\cite{jensen2001subsurface,frisvad2014directional}.
Classical real-time approaches approximate the contribution of surrounding illumination through texture-space diffusion~\cite{deon2007skin}, screen-space and separable filtering~\cite{jimenez2009screenspace,jimenez2015separable}, light-space gathering~\cite{dachsbacher2003translucent}, and image-space irradiance splatting~\cite{shah2009imagespace,chen2012multiresolution}.
Despite their different representations, these methods share the principle of aggregating illumination over a surface region rather than evaluating only a local response.

Recent work learns compact scattering representations~\cite{tg2024neuralsss} and incorporates translucent appearance into Gaussian reconstruction~\cite{dihlmann2024sssgs,zheng2026ssdgs,araneda2026diamondsss}.
SSS-GS~\cite{dihlmann2024sssgs} and SSD-GS~\cite{zheng2026ssdgs} extend Gaussian appearance with learned scattering responses and diffusion-based modeling, respectively.
Building on light-space aggregation, our framework learns light-transport relationships between regional illumination and local appearance, capturing non-local scattering within the Gaussian representation for translucent effects. This supports accurate reconstruction and appearance adjustment of translucent objects.

%% file: sec_3_method.tex
\section{Method}
\label{sec:method}
\input{fig_overview.tex}

\subsection{Preliminaries}
\label{sec:preliminaries}
\paragraph{3D Gaussian Splatting.}
3DGS~\cite{kerbl2023gaussians} represents a scene using anisotropic Gaussians with centers $\boldsymbol\mu_k$, covariances $\boldsymbol\Sigma_k$, and opacities $o_k$.
The kernel of each primitive is
\begin{equation}
 G_k(\mathbf x)=\exp\!\left[-\tfrac12
 (\mathbf x-\boldsymbol\mu_k)^\top\boldsymbol\Sigma_k^{-1}
 (\mathbf x-\boldsymbol\mu_k)\right].
\end{equation}
Projecting these primitives onto the image plane yields footprints whose front-to-back compositing gives
\begin{equation}
 \mathbf C(p)=\sum_k T_k(p)a_k(p)\mathbf c_k,
 \qquad T_k(p)=\prod_{j<k}(1-a_j(p)).
 \label{eq:gs_compositing}
\end{equation}
Here $a_k$ combines opacity with the projected footprint, and $\mathbf c_k$ is the primitive color, conventionally parameterized by spherical harmonics for view dependence.
This representation supports differentiable rendering, but its standard color parameterization does not explicitly respond to changes in illumination.

\paragraph{BSSRDF.}
For translucent materials, light can enter at one surface point $\mathbf x_i$ and leave at another point $\mathbf x_o$. The BSSRDF $\mathcal S$~\cite{jensen2001subsurface} relates light entering and leaving at different points.
For an object that does not emit light, the light leaving $\mathbf x_o$ toward the viewing direction $\boldsymbol\omega_o$ is given by
\begin{equation}
\begin{aligned}
 L_o(\mathbf x_o,\boldsymbol\omega_o)
 &=\int_{\mathcal A}\int_{\Omega^+(\mathbf x_i)}
 \mathcal S(\mathbf x_i,\boldsymbol\omega_i;
             \mathbf x_o,\boldsymbol\omega_o)\\[-2pt]
 &\quad\cdot L_i(\mathbf x_i,\boldsymbol\omega_i)
 (\mathbf n_i^\top\boldsymbol\omega_i)
 \,\mathrm d\boldsymbol\omega_i\,\mathrm dA(\mathbf x_i),
 \label{eq:bssrdf}
\end{aligned}
\end{equation}
where $L_i$ is incident radiance, $\mathbf n_i$ is the surface normal, $\boldsymbol\omega_i$ is the incident light direction, and $\Omega^+(\mathbf x_i)$ is the outward hemisphere.
This motivates extending local Gaussian evaluation to non-local scattering, where illumination from multiple Gaussians across the surface jointly contributes to the appearance at a target surface point.

\subsection{Overview}
\label{sec:decomposed_shading}

The goal of NNS-GS is relighting translucent objects with high quality and supporting plausible scattering adjustment. To this end, NNS-GS introduces two complementary modules built on a shared Gaussian representation with learned latent codes, as illustrated in \cref{fig:overview}. The non-local scattering module (\cref{sec:scattering}) accounts for illumination contributions from other surface locations, enabling more accurate translucent appearance with light space and view space.
The geometry-guided reflection module (\cref{sec:local_reflection}) provides the diffuse illumination cues needed for scattering while using geometric guidance to improve fine specular and shadow details.
We train these modules progressively (\cref{sec:optimization}), using a fixed anchor to separate specular reflection from the diffuse--scattering base.

We express the appearance at a shading point as
\begin{equation}
\begin{aligned}
 L_o&=(\mathbf 1-\mathbf w)\,L_d
                +\mathbf w\,L_{\mathrm{sss}}+L_s,\\
 L_d&=\mathbf k_d\,
L_i(\mathbf x_i,\boldsymbol\omega_i)
\max(0,\mathbf n_i^\top\boldsymbol\omega_i).
\end{aligned}
 \label{eq:base_mix}
\end{equation}
Here $L_d$, $L_{\mathrm{sss}}$, and $L_s$ denote diffuse reflection, subsurface scattering, and specular reflection, respectively.
We denote the diffuse--scattering base by $L_{\mathrm{base}}=(\mathbf 1-\mathbf w)L_d+\mathbf wL_{\mathrm{sss}}$, so that $L_o=L_{\mathrm{base}}+L_s$.
The diffuse component follows a Lambertian form, while scattering and specular reflection are learned by separate networks. The attenuation weight $\mathbf w$ allows editing the balance between scattering and diffuse reflection while keeping specular reflection separate.

\subsection{Dual-Space Non-Local Neural Scattering}
\label{sec:scattering}
\input{fig_local_vs_neighbor_scattering.tex}

The scattering at a shading point comes from light that enters the object at multiple surface locations and travels through the material, which local cues alone cannot capture under new lighting (\cref{fig:local_vs_neighbor_scattering}).
Learning a full transport model involves high-dimensional spatial and angular dependencies, and evaluating it requires aggregating light contributions from the whole surface at substantial cost~\cite{tg2024neuralsss}. However, under point lighting, light-space sampling provides a compact set of illumination cues from light-visible surface regions~\cite{dachsbacher2003translucent}.
This motivates us to propose a non-local neural scattering model that uses light-space cues to learn the prediction of scattered appearance in view space, forming an efficient scattering evaluation.

Our scattering prediction proceeds in two stages (\cref{fig:scattering_combined}). First, we aggregate illumination cues as non-local features in light space, including diffuse responses and distance from light-visible surface samples. Second, a neural model combines these non-local cues with local Gaussian features to predict scattered appearance in view space.

\paragraph{Light-space feature aggregation.}
\input{fig_scattering_combined.tex}
In light space, this stage aggregates non-local illumination cues and their spatial relationships to each shading point $\mathbf P$. To describe illumination across the light-visible surface, we use the Lambertian diffuse response, which shares the incoming-light and cosine terms of the BSSRDF in \cref{eq:bssrdf}.
We first splat the Gaussians from the light viewpoint to jointly obtain maps of depth, 3D surface positions, and latent codes, using light-space splatting~\cite{bi2024gs3}.
Then, the shading point is projected into these maps, and a fixed ring-based pattern samples around its projection. Samples in this pattern are denser on the inner rings to emphasize nearby contributions and sparser on the outer rings to cover farther regions.
Each valid sample provides a diffuse response $\mathbf q_j$ through the diffuse network and a Euclidean distance $d_j$ from its stored 3D position to $\mathbf P$.
The $K=32$ samples form the non-local descriptor $\mathcal N_L(\mathbf P)=[\mathbf q_1,d_1,\ldots,\mathbf q_K,d_K]$ in a fixed sampling order, describing the illumination and spatial relationships within the contributing surface region.

\paragraph{Learned scattering network.}
In view space, a scattering network $f_{\mathrm{sss}}$ is used to implicitly learn how illumination at the sampled surface locations jointly contributes to the shading point $\mathbf P$.
To describe this relationship, $f_{\mathrm{sss}}$ takes the non-local descriptor $\mathcal N_L(\mathbf P)$ together with the local Gaussian latent code $z$ and incidence angle $\theta$.
Here $z$ encodes local material and geometry, and $\theta$ is the angle between the direction toward the light and the proxy normal extracted from the learned local frame (\cref{sec:local_reflection}). A distance-dependent modulation $\mathbf A$ attenuates the learned transport response to give the scattering component,
\begin{equation}
 L_{\mathrm{sss}}=
 \mathbf A\, f_{\mathrm{sss}}
 \big(\mathcal N_L(\mathbf P),z,\theta\big).
 \label{eq:neural_scattering}
\end{equation}
To support plausible adjustments to scattering appearance, we introduce a nonnegative parameter $s$ in $A$,
\begin{equation}
 A=\frac{1}{|\mathcal V|}
       \sum_{j\in\mathcal V}\exp(-s d_j),
 \label{eq:regional_attenuation}
\end{equation}
where $\mathcal V$ contains valid samples. Within the exponential attenuation factors, $s$ acts as a relative effective-extinction scale (see Sec.~A of the supplementary material for the rationale). This physically inspired parameter modulates scattering appearance while keeping the learned transport response fixed. By default, $s$ also scales the per-channel optical-thickness proxy $\boldsymbol\tau$, giving the scattering weight:
\begin{equation}
 \mathbf w=\exp(-s\boldsymbol\tau),
 \label{eq:extinction_blend}
\end{equation}
where $\boldsymbol\tau>0$ is learned from $z$. This gives a single control for both extinction attenuation and the diffuse--scattering blend in \cref{eq:base_mix}. Increasing $s$ attenuates scattering more strongly and increases the complementary diffuse weight $\mathbf 1-\mathbf w$, shifting the appearance towards diffuse reflection.

\subsection{Geometry-Guided Neural Reflection}
\label{sec:local_reflection}
Alongside non-local scattering, relighting requires an accurate estimate of the light reflected at the local surface point. The challenges arise: highlights change sharply with the viewing direction, and cast shadows can have sharp boundaries even on a smooth surface. MLPs tend to learn smooth changes more readily than abrupt variations~\cite{rahaman2019spectral}, making these fine details harder to recover. We therefore propose a geometry-guided neural reflection module, using additional geometry cues to guide the prediction of these details.

The module uses two branches for diffuse and specular reflection, sharing a Gaussian latent code $z$ that encodes material and geometry (\cref{fig:reflection_network}). In this network, a small MLP decodes a local frame from $z$, providing both a proxy normal for diffuse reflection and a surface-relative local coordinate system for specular prediction. Finally, a shared shadow modulation factor adjusts both outputs.

Specifically, the diffuse branch takes $z$ together with light-intensity and depth-based shadow cues~\cite{fan2025rng}, to decode the diffuse coefficient. With the proxy normal, the diffuse response is then computed using the Lambertian form, combining the learned coefficient with the incoming light. On the other hand, the specular branch directly predicts highlight reflection from $z$ through an MLP rather than evaluating an analytic BRDF~\cite{bi2024gs3,zheng2026ssdgs}. The light and view directions are transformed into the learned local frame before entering this branch, reducing the burden on the MLP to learn spatially varying directional rotations~\cite{zeltner2024neuralappearance}. Their half-vector $h$, defined as the normalized sum of these two directions, is also transformed into the same frame as inputs.

\input{fig_reflection_network.tex}
To improve shadow quality, we propose a shadow modulation module to adjust the output of diffuse and specular reflection.
The module uses the difference between the shading point's light-space depth and the rendered depth at its projection as a shadow cue. A small MLP therefore takes the latent code and a low-frequency encoding of this cue to predict a shared modulation factor, which is applied to both branches. Note that this learned adjustment can attenuate or amplify reflection but is not a physical visibility term.

\subsection{Anchor-Based Progressive Training}
\label{sec:optimization}
Under a standard training scheme~\cite{fan2025rng}, the specular branch can absorb diffuse and scattering contributions. Reconstruction may remain accurate, but these contributions no longer respond to adjustments of scattering attenuation.
We therefore use anchor-based progressive training to separate $L_{\mathrm{base}}$ from $L_{s}$. It learns diffuse reflection first, introduces scattering, and saves the combined response $L_{\mathrm{base}}$ as a supervision anchor before training the specular.
We adopt a two-stage training scheme, using forward splatting for geometry and deferred splatting~\cite{guennebaud2004deferred} for appearance.

The deferred stage uses 70k iterations in total.
During the first 10k iterations, only diffuse reflection is learned, providing a stable illumination basis for scattering.
The next 20k iterations introduce scattering and jointly refine the diffuse--scattering base.
At iteration 30k, the rendered base for each training view $i$ is stored as a fixed anchor $\mathbf H_i$.
The remaining 40k iterations introduce specular reflection and supervise the base and specular components separately.
The base is constrained to remain close to $\mathbf H_i$, while the specular target is the positive residual $\mathbf R_i=[\mathbf Y_i-\mathbf H_i]_+$ between the observed foreground image $\mathbf Y_i$ and the anchor.
Here $[\cdot]_+$ retains positive values and sets negative values to zero.
The anchors remain fixed during this final phase, giving the two components distinct supervision targets.

\paragraph{Training losses.}
The loss terms vary in the training stages. Diffuse learning uses $\mathcal L_{\mathrm{rec}}$, combining $L_1$ and SSIM-based reconstruction terms. Scattering adds a total-variation loss $\mathcal L_{\mathrm{TV}}$ for spatial smoothness. Specular learning retains both and adds separate supervision,
\begin{equation}
\begin{aligned}
 \mathcal L={}&\mathcal L_{\mathrm{rec}}
       +\lambda_{\mathrm{TV}}\mathcal L_{\mathrm{TV}}\\
       &+\lambda_{\mathrm{anchor}}\mathcal L_{\mathrm{anchor}}
       +\lambda_{\mathrm{spec}}\mathcal L_{\mathrm{spec}},
 \label{eq:training_objective}
\end{aligned}
\end{equation}
where $\mathcal L_{\mathrm{anchor}}$ is the foreground $L_1$ difference between the base and its anchor. Since specular reflection is sparse, uniform supervision can underemphasize fine detail. Accordingly, $\mathcal L_{\mathrm{spec}}$ combines foreground $L_1$ error against $\mathbf R_i$ with a normalized error weighted by residual intensity. Using the maximum residual RGB channel also emphasizes colored specular responses, while the full-foreground term constrains unwanted responses elsewhere.

\input{fig_main_comparison.tex}

%% file: fig_overview.tex
\begin{figure*}[t]
    \centering
    \includegraphics[width=\textwidth]{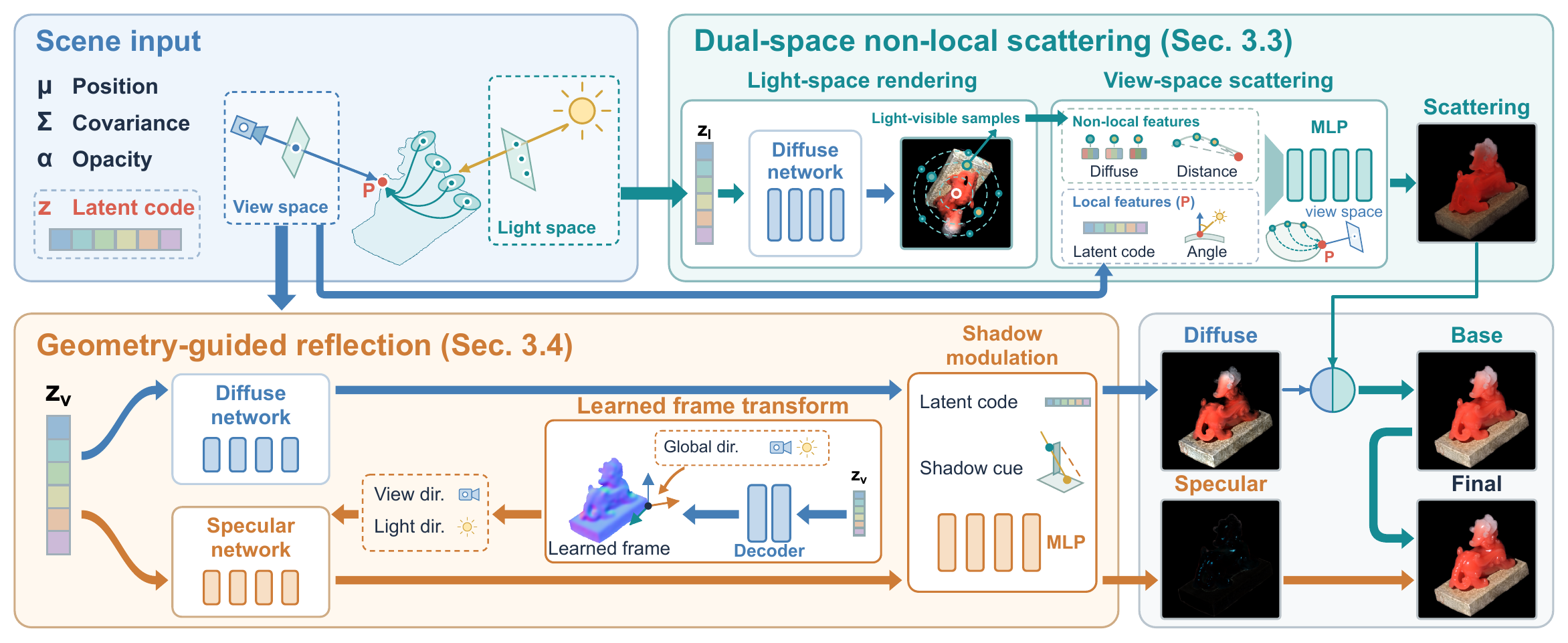}
    \caption{Overview of the NNS-GS framework. Our framework comprises two modules for non-local scattering and detailed local reflection. The scattering module collects illumination cues from light-visible surfaces in light space and learns their contributions to the point $\mathbf P$ in view space with both non-local and local features. The reflection module uses two branches to supply diffuse cues and predict specular reflection, using learned local frames and shadow modulation to refine appearance details. All components form the final image. Here \(z_l\) and \(z_v\) denote latent codes rasterized in light and view space, respectively. The diffuse network is shared across the two spaces.}
    \label{fig:overview}
\end{figure*}

%% file: fig_local_vs_neighbor_scattering.tex
\begin{figure}[t]
    \centering
    \includegraphics[width=\columnwidth]{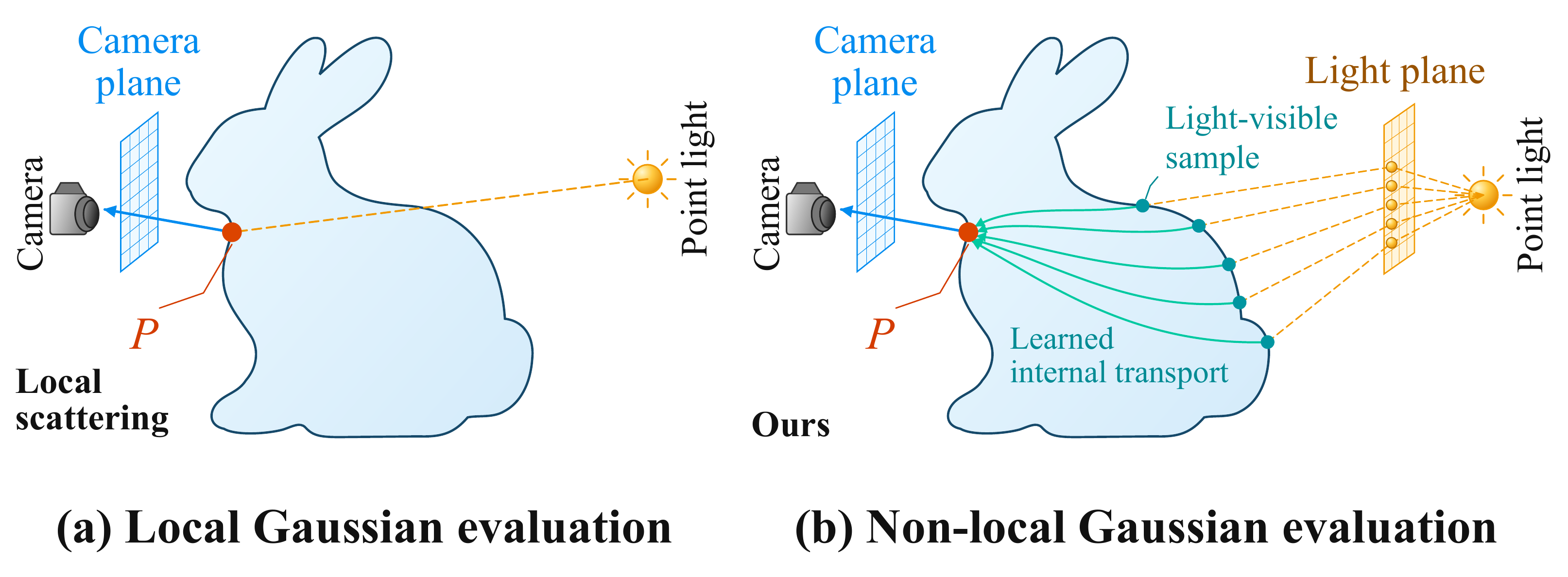}
    \caption{Local versus non-local scattering evaluation. (a) Local methods~\cite{dihlmann2024sssgs,zheng2026ssdgs} use information from an individual Gaussian to evaluate scattering at a surface point. (b) Our method gathers illumination from multiple Gaussians on the light-visible surface and learns how they jointly contribute to the target point.}
    \label{fig:local_vs_neighbor_scattering}
\end{figure}

%% file: fig_scattering_combined.tex
\begin{figure}[t]
    \centering
    \includegraphics[width=\columnwidth]{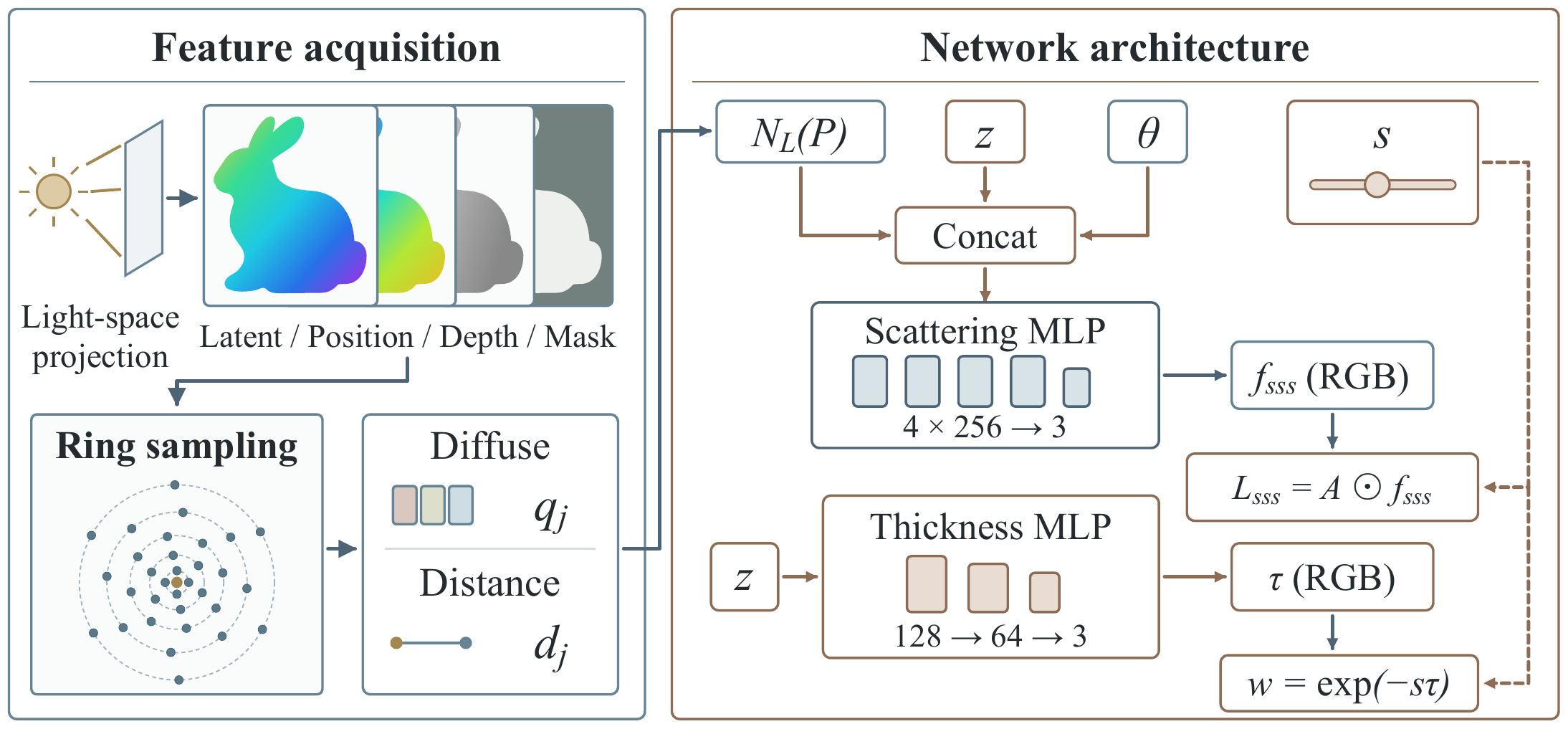}
    \caption{Non-local neural scattering. Feature acquisition combines light-space projection with ring-based sampling to obtain non-local diffuse responses and distances. The network combines these features with the local latent code and incidence angle to predict scattering. The editing scale controls distance attenuation and diffuse--scattering mixing.}
    \label{fig:scattering_combined}
    \label{fig:light_space_sampling}
    \label{fig:scattering_network}
\end{figure}

%% file: fig_reflection_network.tex
\begin{figure}[t]
    \centering
    \includegraphics[width=\columnwidth]{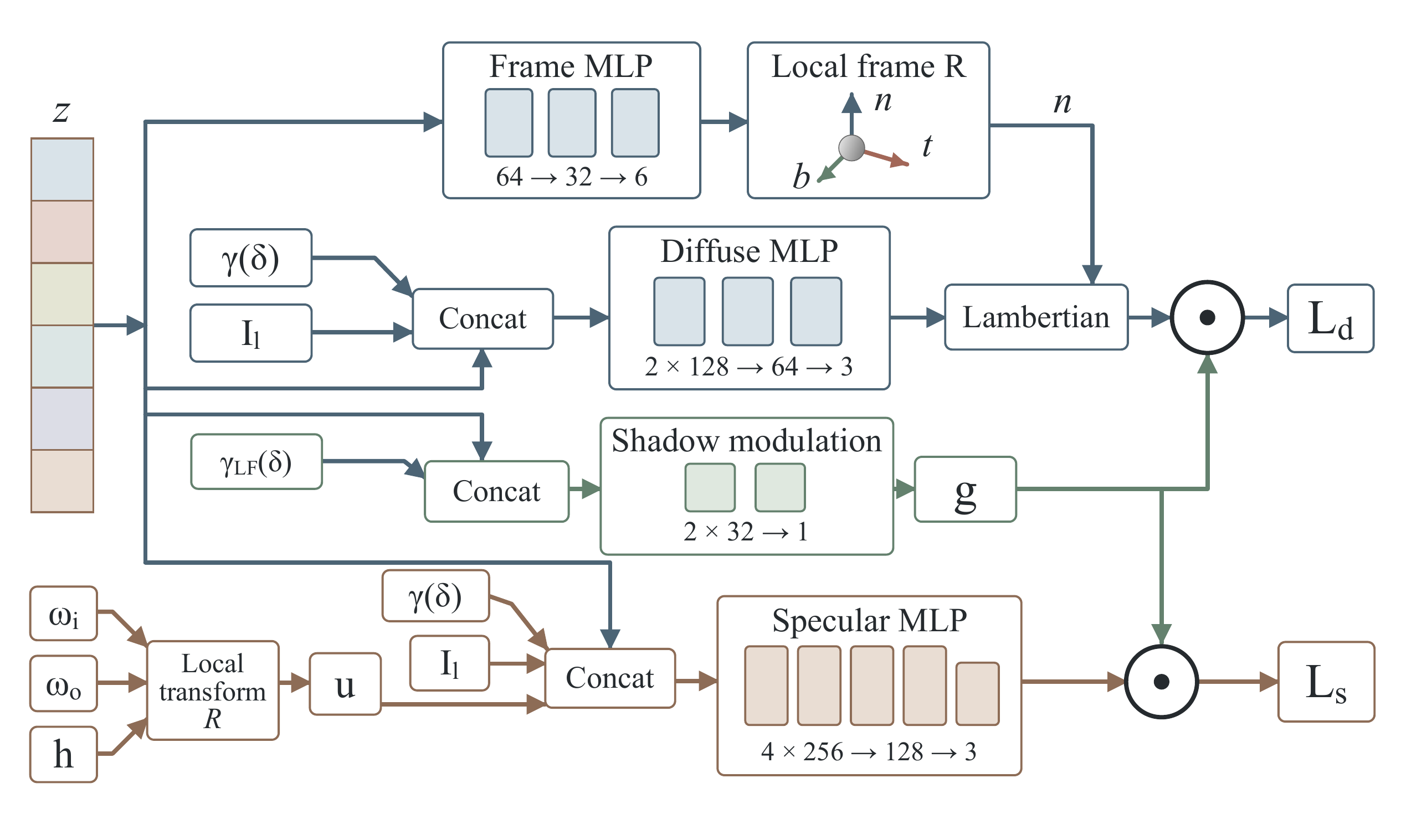}
    \caption{Geometry-guided neural reflection. Two 3D vectors from the 6D output are orthonormalized into a normal and tangent, whose cross product completes the local frame. The frame supplies the diffuse normal and local specular directions, while $g$ modulates both branches. Here $\delta$ denotes the light-space depth difference, $\gamma$ and $\gamma_{\mathrm{LF}}$ its full and low-frequency encodings, $I_l$ light intensity, and $\mathbf u$ local light, view, and half-vector directions.}
    \label{fig:reflection_network}
\end{figure}

%% file: fig_main_comparison.tex
\begin{figure*}[t]
    \centering
    \includegraphics[width=\textwidth]{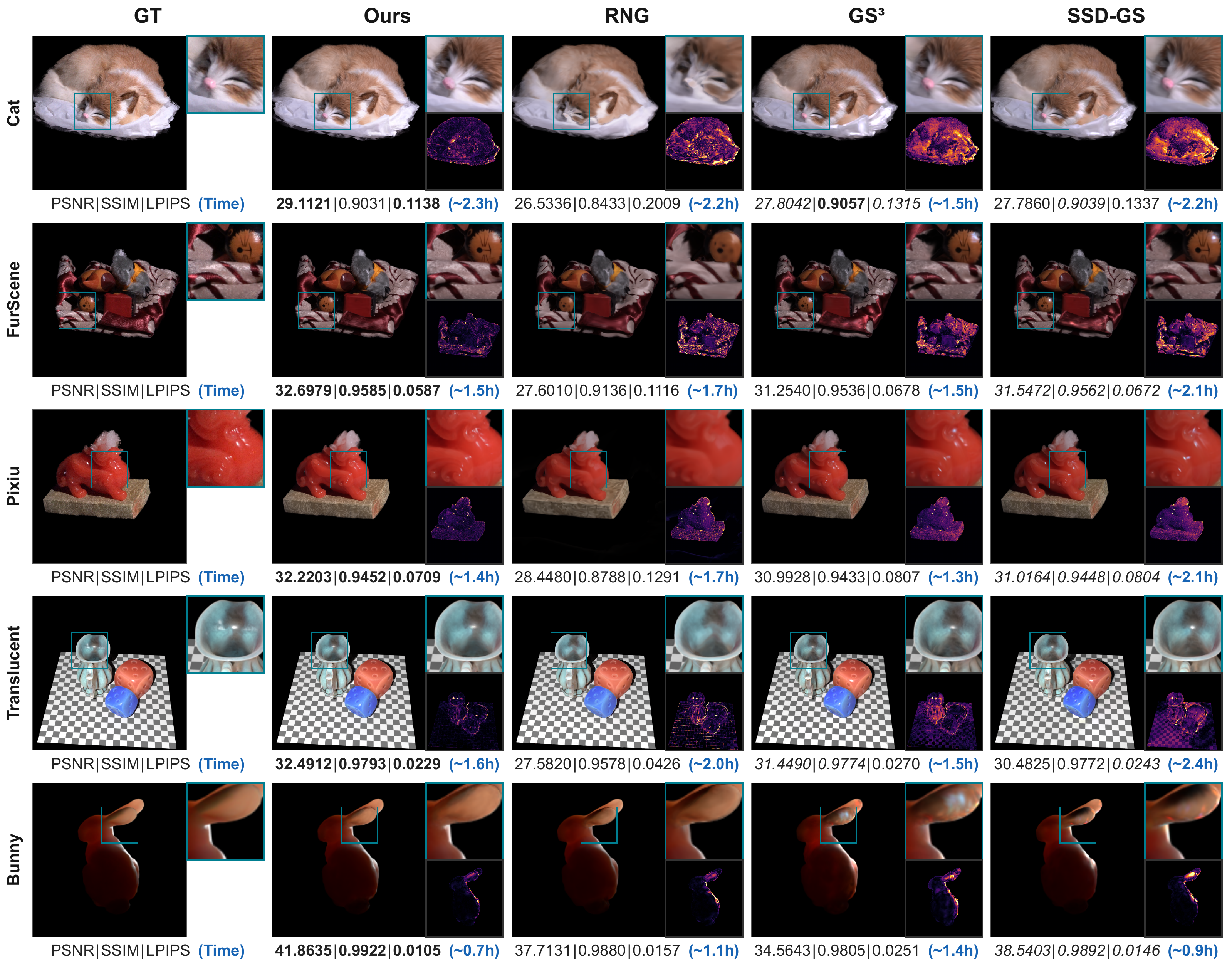}
    \caption{Qualitative comparison among ground truth, ours, RNG~\cite{fan2025rng}, GS\textsuperscript{3}~\cite{bi2024gs3}, and SSD-GS~\cite{zheng2026ssdgs}. The best/second-best results are marked as bold/italic. Numbers report mean PSNR / SSIM / LPIPS (Training Time). More results can be seen in the supplementary material.}
    \label{fig:main_comparison}
\end{figure*}

%% file: sec_4_experiments.tex
\section{Experiments}
\label{sec:experiments}

\input{fig_results_summary.tex}

\subsection{Experimental Setup}
\label{sec:experimental_setup}

\paragraph{Implementation.}
We implement NNS-GS in PyTorch on an NVIDIA GeForce RTX 4090 D, using Adam~\cite{kingma2015adam} and $K=32$ ring-based samples per shading point.
Forward splatting and deferred appearance training use 30k and 70k iterations, respectively.
Training typically takes 1--2 hours per scene, with rendering at about 60 FPS on the same GPU.
For more implementation details, please refer to the supplementary material (Sec.~A).

\paragraph{Datasets.}
We evaluate on datasets from SSS-GS~\cite{dihlmann2024sssgs}, GS\textsuperscript{3}~\cite{bi2024gs3}, and NRHints~\cite{zeng2023nrhints}, covering translucent and general objects under varying point lighting. We follow the official train/test splits and evaluate on the test sets. For SSS-GS datasets, we additionally evaluate on test images with unseen light positions to assess relighting generalization. All datasets use a consistent black background.

\paragraph{Baselines.}
We compare against RNG~\cite{fan2025rng}, SSS-GS~\cite{dihlmann2024sssgs}, GS\textsuperscript{3}~\cite{bi2024gs3}, and SSD-GS~\cite{zheng2026ssdgs}.
RNG and GS\textsuperscript{3} represent neural and spatial--angular Gaussian methods for relighting, while SSS-GS and SSD-GS explicitly address SSS. We additionally compare with GI-GS~\cite{chen2025gigs} and GS-IR~\cite{liang2024gsir} to evaluate our method under environment lighting. All baselines are evaluated using released official implementations.

\subsection{Validation}
\label{sec:quantitative_qualitative}

\Cref{tab:sssgs_test} and Sec. B of the supplementary material show the PSNR, SSIM, and LPIPS across all three datasets. NNS-GS achieves the best mean score for each metric on every dataset.
The gains are particularly pronounced on the translucent scenes from SSS-GS, with mean PSNR improvements over RNG~\cite{fan2025rng} of 2.47~dB on synthetic scenes and 0.81~dB on real scenes.
\Cref{fig:main_comparison} further compares visual results on synthetic and real-world scenes. NNS-GS more faithfully reproduces scattering appearance across different materials. On Bunny, it better captures the soft scattering-induced brightness variations across the back-lit ear.
The benefits extend to materials whose appearance also depends on non-local light transport, with better-preserved fur detail on Cat and FurScene.
Compared with RNG, which also uses a neural BRDF, NNS-GS reproduces finer specular reflections on Pixiu and more accurate shadows on FurScene.
Beyond point lighting, NNS-GS also outperforms the environment-relighting Gaussian methods GI-GS~\cite{chen2025gigs} and GS-IR~\cite{liang2024gsir}, achieving the best score on evaluated scenes (Sec.~B of the supplementary material).

To examine the scattering controls, we construct synthetic reference scenes in Blender~\cite{blender2026lts} following SSS-GS~\cite{dihlmann2024sssgs}, and isolate extinction and blending to match the baseline parameterizations.
For extinction-related editing, we vary $A$ with $w$ fixed and compare with SSD-GS~\cite{zheng2026ssdgs}, which uses explicit optical coefficients. For blending editing, we vary the original $w$ with $L_{\mathrm{sss}}$ fixed and compare with SSS-GS~\cite{dihlmann2024sssgs}, which provides an explicit blending weight. \Cref{fig:components_scales} shows the default and edited results. Our method achieves higher PSNR in the illustrated settings, indicating closer agreement with the edited references.

\subsection{Ablation Studies}
\label{sec:ablations}
\input{sec_4_ablation_table.tex}

\input{fig_ablation_comparison.tex}

\input{fig_component_supervision_editing.tex}

\paragraph{Scattering Module.}
Our scattering module combines local view-space features with non-local illumination cues gathered in light space.
Removing the module reduces mean PSNR by 1.73~dB across the three scenes (\cref{tab:component_ablations}) and produces an overly sharp shading transition (\cref{fig:ablation_comparison}).
The scattering module with only local view-space features, including the diffuse response, improves mean PSNR over the no-scattering variant but underperforms the full model. Without non-local features, this variant also fails to accurately reproduce the light transmitted through Bunny's ear (\cref{fig:ablation_comparison}), which supports the benefit of incorporating non-local illumination cues beyond local scattering prediction.

\paragraph{Frame transform.}
Our specular branch expresses the light and view directions, together with the half-vector $h$, in a learned local frame.
Removing both the frame transformation and $h$ reduces reconstruction quality, with less faithful fine-scale specular detail on Pixiu (\cref{fig:ablation_comparison}).
Omitting only $h$ while retaining the transformation also degrades performance (see supplementary material).

\paragraph{Shadow modulation.}
Our shadow modulation uses light-space depth cues to jointly adjust diffuse and specular reflection.
Without this modulation, Translucent exhibits more irregular shadow boundaries and uneven ground-plane shading (\cref{fig:ablation_comparison}), accompanied by lower reconstruction quality.
Its effect is smaller on Pixiu and Bunny (\cref{tab:component_ablations}), where these artifacts are less prominent.

\paragraph{Anchor supervision.}
During progressive training, a fixed anchor guides the separate supervision of the diffuse--scattering base and specular branch.
Without anchor-based separate supervision, the specular branch absorbs base color and leaves the diffuse--scattering component darker (\cref{fig:component_supervision_editing}). This component mixing is problematic for scattering controls, as the contributions no longer respond as intended, increasing the discrepancy between the edited result and the reference.

\subsection{Limitations}
A limitation of our method is its reduced advantage on metallic scenes such as Drums. Their appearance is dominated by local reflection, leaving limited benefit from the non-local scattering module. Moreover, their sharp, strongly anisotropic specular reflections remain difficult for MLPs to represent accurately, even with our geometry-guided reflection model. Further details are provided in the supplementary discussion (Sec.~C).

%% file: fig_results_summary.tex
\begin{figure*}[!t]
\centering
\begingroup
\renewenvironment{table*}[1][]{\par\begingroup\captionsetup{type=table}}{\par\endgroup}
\input{sec_4_test_tables.tex}

\endgroup
\medskip
\begingroup
\renewenvironment{figure*}[1][]{\par\captionsetup{type=figure}}{\par}
\input{fig_components_scales.tex}
\endgroup
\end{figure*}

%% file: sec_4_test_tables.tex
\begin{table*}[t]
\centering
\caption{Quantitative comparison on synthetic and real-world scenes from SSS-GS~\cite{dihlmann2024sssgs}. All denotes the full test split and relight denotes its subset with light positions absent from training. Red and orange cells indicate the best and second-best results, respectively. }
\label{tab:sssgs_test}
\begingroup
\footnotesize
\color{black}
\setlength{\tabcolsep}{1pt}
\renewcommand{\arraystretch}{1.0}
\begin{tabular*}{\textwidth}{@{\extracolsep{\fill}}l*{15}{r}@{}}
\toprule
Scene & \multicolumn{3}{c}{GS\textsuperscript{3}} & \multicolumn{3}{c}{SSD-GS} & \multicolumn{3}{c}{RNG} & \multicolumn{3}{c}{SSS-GS} & \multicolumn{3}{c}{Ours} \\
\cmidrule(lr){2-4}\cmidrule(lr){5-7}\cmidrule(lr){8-10}\cmidrule(lr){11-13}\cmidrule(lr){14-16}
 & PSNR $\uparrow$ & SSIM $\uparrow$ & LPIPS $\downarrow$ & PSNR $\uparrow$ & SSIM $\uparrow$ & LPIPS $\downarrow$ & PSNR $\uparrow$ & SSIM $\uparrow$ & LPIPS $\downarrow$ & PSNR $\uparrow$ & SSIM $\uparrow$ & LPIPS $\downarrow$ & PSNR $\uparrow$ & SSIM $\uparrow$ & LPIPS $\downarrow$ \\
\midrule
Bunny & 34.5643 & 0.9805 & 0.0251 & \cellcolor{orange!25}38.5403 & \cellcolor{orange!25}0.9892 & \cellcolor{orange!25}0.0146 & 37.7131 & 0.9880 & 0.0157 & 35.5260 & 0.9785 & 0.0258 & \cellcolor{red!20}41.8635 & \cellcolor{red!20}0.9922 & \cellcolor{red!20}0.0105 \\
Candle & 38.6118 & 0.9886 & 0.0168 & 41.2362 & 0.9929 & 0.0089 & \cellcolor{orange!25}43.2076 & \cellcolor{orange!25}0.9956 & \cellcolor{orange!25}0.0060 & 33.7893 & 0.9763 & 0.0285 & \cellcolor{red!20}45.2061 & \cellcolor{red!20}0.9961 & \cellcolor{red!20}0.0056 \\
Dragon & 35.8287 & 0.9757 & 0.0262 & \cellcolor{orange!25}37.4180 & \cellcolor{red!20}0.9818 & \cellcolor{red!20}0.0210 & 35.2122 & 0.9731 & 0.0311 & 35.1088 & 0.9628 & 0.0402 & \cellcolor{red!20}37.6829 & \cellcolor{orange!25}0.9814 & \cellcolor{orange!25}0.0212 \\
Soap & 38.3513 & 0.9840 & 0.0252 & 38.5117 & 0.9879 & 0.0177 & \cellcolor{orange!25}40.3410 & \cellcolor{orange!25}0.9905 & \cellcolor{red!20}0.0104 & 36.7401 & 0.9761 & 0.0232 & \cellcolor{red!20}42.2419 & \cellcolor{red!20}0.9922 & \cellcolor{orange!25}0.0107 \\
Statue & 36.4952 & 0.9846 & 0.0183 & 35.3636 & 0.9818 & 0.0212 & \cellcolor{orange!25}37.5505 & \cellcolor{orange!25}0.9883 & \cellcolor{orange!25}0.0143 & 35.2588 & 0.9754 & 0.0273 & \cellcolor{red!20}39.3792 & \cellcolor{red!20}0.9896 & \cellcolor{red!20}0.0141 \\
\midrule
Mean (syn., all) & 36.7702 & 0.9827 & 0.0223 & 38.2139 & 0.9867 & 0.0167 & \cellcolor{orange!25}38.8049 & \cellcolor{orange!25}0.9871 & \cellcolor{orange!25}0.0155 & 35.2846 & 0.9738 & 0.0290 & \cellcolor{red!20}41.2747 & \cellcolor{red!20}0.9903 & \cellcolor{red!20}0.0124 \\
Mean (syn., relight) & 35.9780 & 0.9798 & 0.0252 & 37.3340 & 0.9840 & 0.0196 & \cellcolor{orange!25}38.2040 & \cellcolor{orange!25}0.9853 & \cellcolor{orange!25}0.0182 & 34.9780 & 0.9718 & 0.0316 & \cellcolor{red!20}40.3826 & \cellcolor{red!20}0.9882 & \cellcolor{red!20}0.0157 \\
\midrule
clear\_crystal & 34.7519 & 0.9793 & 0.0266 & 35.3086 & \cellcolor{orange!25}0.9798 & \cellcolor{orange!25}0.0225 & \cellcolor{orange!25}35.5002 & 0.9795 & 0.0235 & 34.4781 & 0.9764 & 0.0242 & \cellcolor{red!20}35.9923 & \cellcolor{red!20}0.9825 & \cellcolor{red!20}0.0177 \\
green\_candle & 30.3782 & 0.9575 & 0.0721 & 30.0874 & 0.9567 & 0.0782 & \cellcolor{orange!25}31.5138 & \cellcolor{orange!25}0.9631 & 0.0668 & 29.7595 & 0.9485 & \cellcolor{red!20}0.0630 & \cellcolor{red!20}32.6282 & \cellcolor{red!20}0.9648 & \cellcolor{orange!25}0.0644 \\
joyful\_yell & 30.7976 & 0.9617 & 0.0504 & 29.5301 & 0.9566 & 0.0557 & \cellcolor{orange!25}32.1884 & \cellcolor{orange!25}0.9678 & \cellcolor{orange!25}0.0402 & 31.0713 & 0.9591 & 0.0461 & \cellcolor{red!20}33.0147 & \cellcolor{red!20}0.9711 & \cellcolor{red!20}0.0338 \\
\midrule
Mean (real, all) & 31.9759 & 0.9661 & 0.0497 & 31.6420 & 0.9644 & 0.0522 & \cellcolor{orange!25}33.0675 & \cellcolor{orange!25}0.9701 & \cellcolor{orange!25}0.0435 & 31.7696 & 0.9613 & 0.0444 & \cellcolor{red!20}33.8784 & \cellcolor{red!20}0.9728 & \cellcolor{red!20}0.0387 \\
Mean (real, relight) & 31.9600 & 0.9660 & 0.0499 & 31.6167 & 0.9643 & 0.0524 & \cellcolor{orange!25}33.0467 & \cellcolor{orange!25}0.9700 & \cellcolor{orange!25}0.0437 & 31.7400 & 0.9612 & 0.0446 & \cellcolor{red!20}33.8433 & \cellcolor{red!20}0.9727 & \cellcolor{red!20}0.0389 \\
\bottomrule
\end{tabular*}
\endgroup
\end{table*}

%% file: fig_components_scales.tex
\begin{figure*}[!t]
    \centering
    \includegraphics[width=\textwidth]{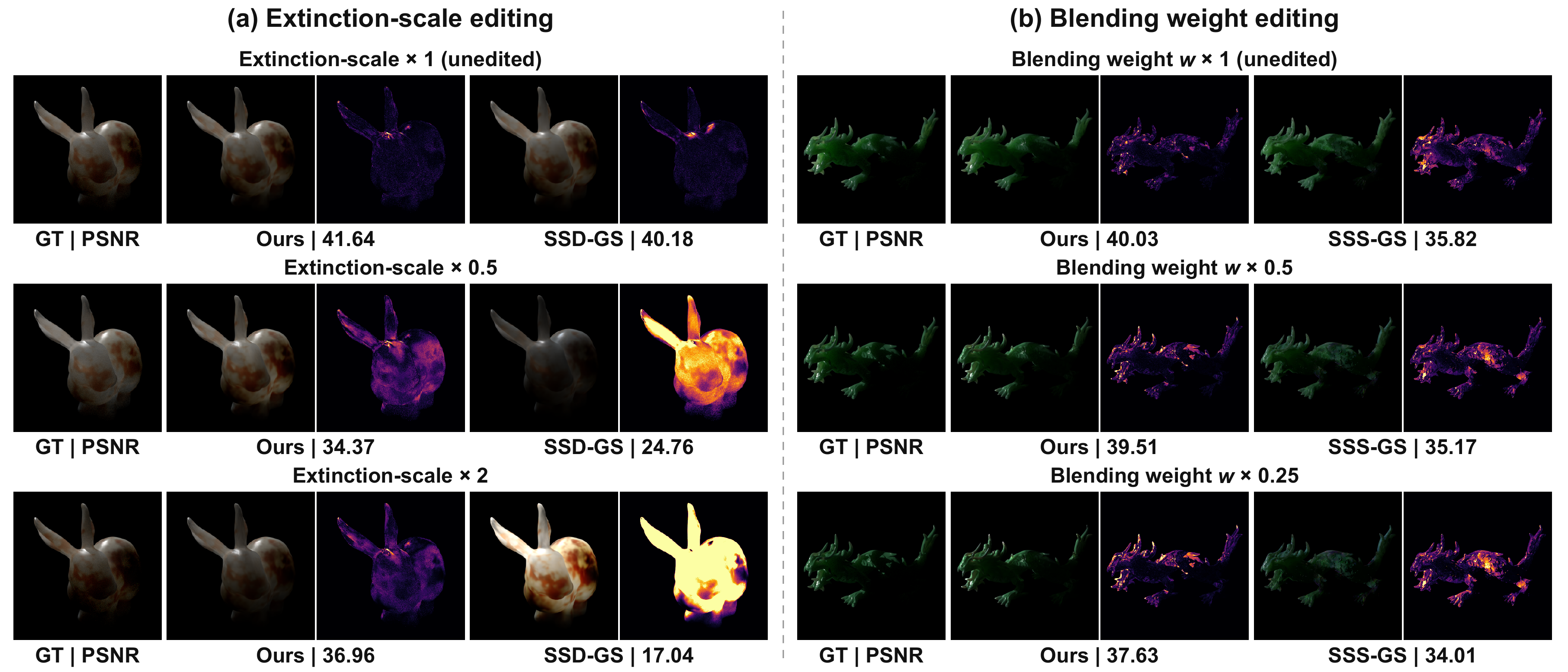}
    \caption{Controlled appearance adjustments on synthetic scenes with a shared factor $a$ ($1$ unedited). (a) With blending fixed, GT scales extinction $\sigma_t$ by $a$, SSD-GS~\cite{zheng2026ssdgs} scales its scattering $\sigma_s$ and absorption $\sigma_a$ ($\sigma_t=\sigma_s+\sigma_a$) by $a$, and ours sets $s=a$ in \cref{eq:regional_attenuation}. (b) With scattering and specular reflection fixed, GT, ours, and SSS-GS scale their original blending weight to $w$ (unedited), $0.5w$, and $0.25w$.}
    \label{fig:components_scales}
\end{figure*}

%% file: sec_4_ablation_table.tex
\begin{table}[tbp]
\centering
\caption{Ablation study of the proposed components. View-space only retains scattering but uses only the shading point in the camera view, without light-space features (\cref{sec:scattering}). frame trans. denotes the learned frame transform (\cref{sec:local_reflection}), shadow mod. denotes shadow modulation (\cref{sec:local_reflection}), and anchor sup. denotes anchor supervision (\cref{sec:optimization}). Best results are bold. Note that even when quality changes little, removing the anchor can increase the discrepancy from the edited reference (\cref{fig:component_supervision_editing}).}
\label{tab:component_ablations}
\begingroup
\footnotesize
\setlength{\tabcolsep}{3pt}
\renewcommand{\arraystretch}{1.08}
\begin{tabular*}{\columnwidth}{@{\extracolsep{\fill}}llrrr@{}}
\toprule
Scene & Configuration & PSNR $\uparrow$ & SSIM $\uparrow$ & LPIPS $\downarrow$ \\
\midrule
Translucent & Full / K32 & \textbf{32.4912} & \textbf{0.9793} & \textbf{0.0229} \\
 & w/o scattering & 30.4112 & 0.9766 & 0.0258 \\
 & View only / K1 & 31.4258 & 0.9763 & 0.0268 \\
 & w/o coord. trans. & 31.3188 & 0.9758 & 0.0262 \\
 & w/o shadow mod. & 29.1219 & 0.9708 & 0.0296 \\
 & w/o anchor & 32.3884 & 0.9768 & 0.0255 \\
\midrule
Pixiu & Full / K32 & \textbf{32.2203} & \textbf{0.9452} & \textbf{0.0709} \\
 & w/o scattering & 31.2919 & 0.9434 & 0.0752 \\
 & View only / K1 & 31.2143 & 0.9431 & 0.0729 \\
 & w/o coord. trans. & 31.8920 & 0.9440 & 0.0739 \\
 & w/o shadow mod. & 32.1021 & 0.9447 & 0.0717 \\
 & w/o anchor & 32.1058 & 0.9447 & 0.0729 \\
\midrule
Bunny & Full / K32 & \textbf{41.8635} & \textbf{0.9922} & \textbf{0.0105} \\
 & w/o scattering & 39.6936 & 0.9897 & 0.0139 \\
 & View only / K1 & 40.1996 & 0.9899 & 0.0137 \\
 & w/o coord. trans. & 41.3181 & 0.9916 & 0.0113 \\
 & w/o shadow mod. & 41.7247 & 0.9918 & 0.0112 \\
 & w/o anchor & 41.6997 & 0.9915 & 0.0119 \\
\bottomrule
\end{tabular*}
\endgroup
\end{table}

%% file: fig_ablation_comparison.tex
\begin{figure}[t]
    \centering
    \includegraphics[width=\columnwidth]{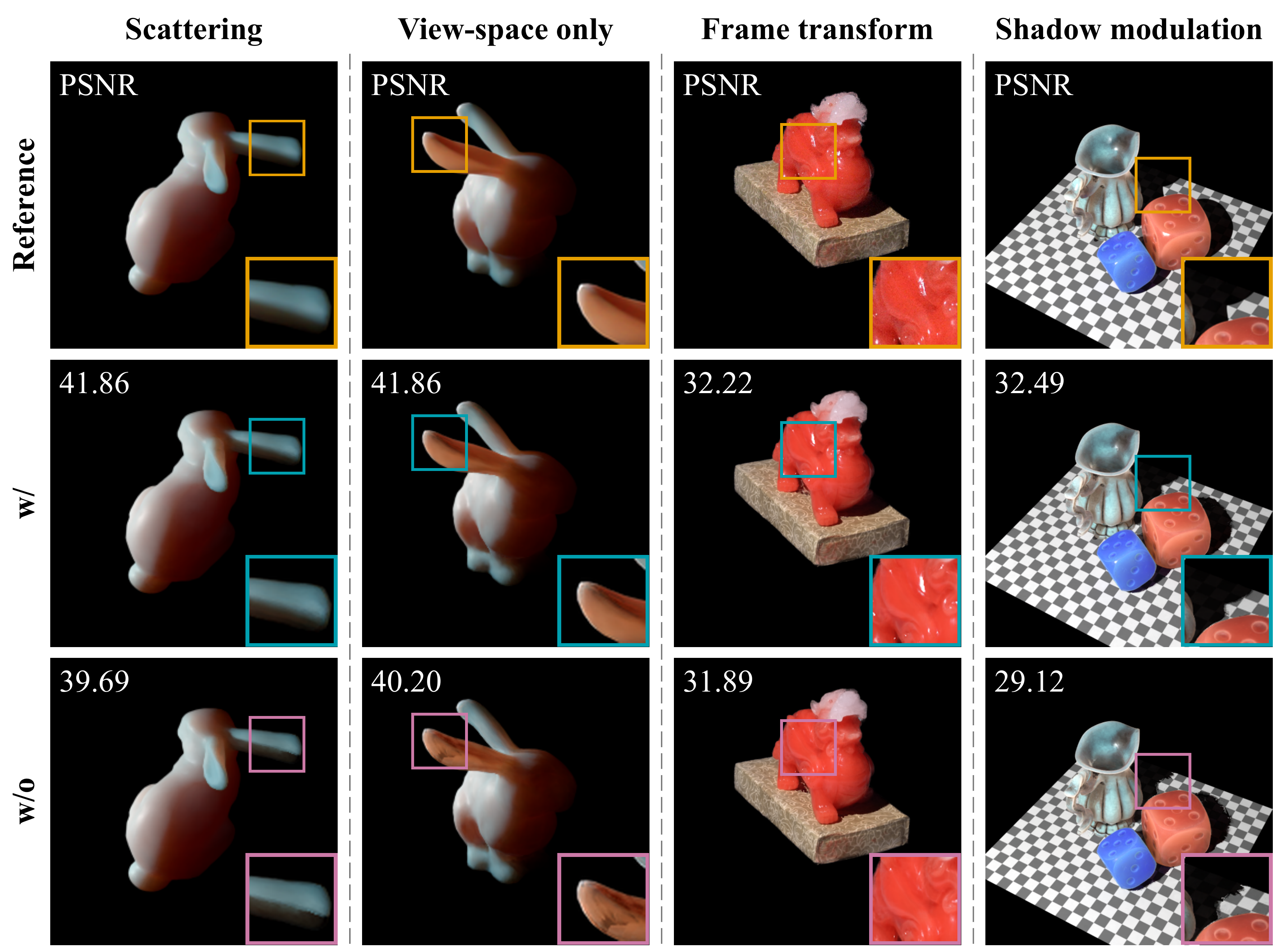}
    \caption{Qualitative ablations showing the effects of the scattering module, view-space-only evaluation, frame transform, and shadow modulation, from left to right. Values denote mean PSNR.}
    \label{fig:ablation_comparison}
\end{figure}

%% file: fig_component_supervision_editing.tex
\begin{figure}[t]
    \centering
    \includegraphics[width=\columnwidth]{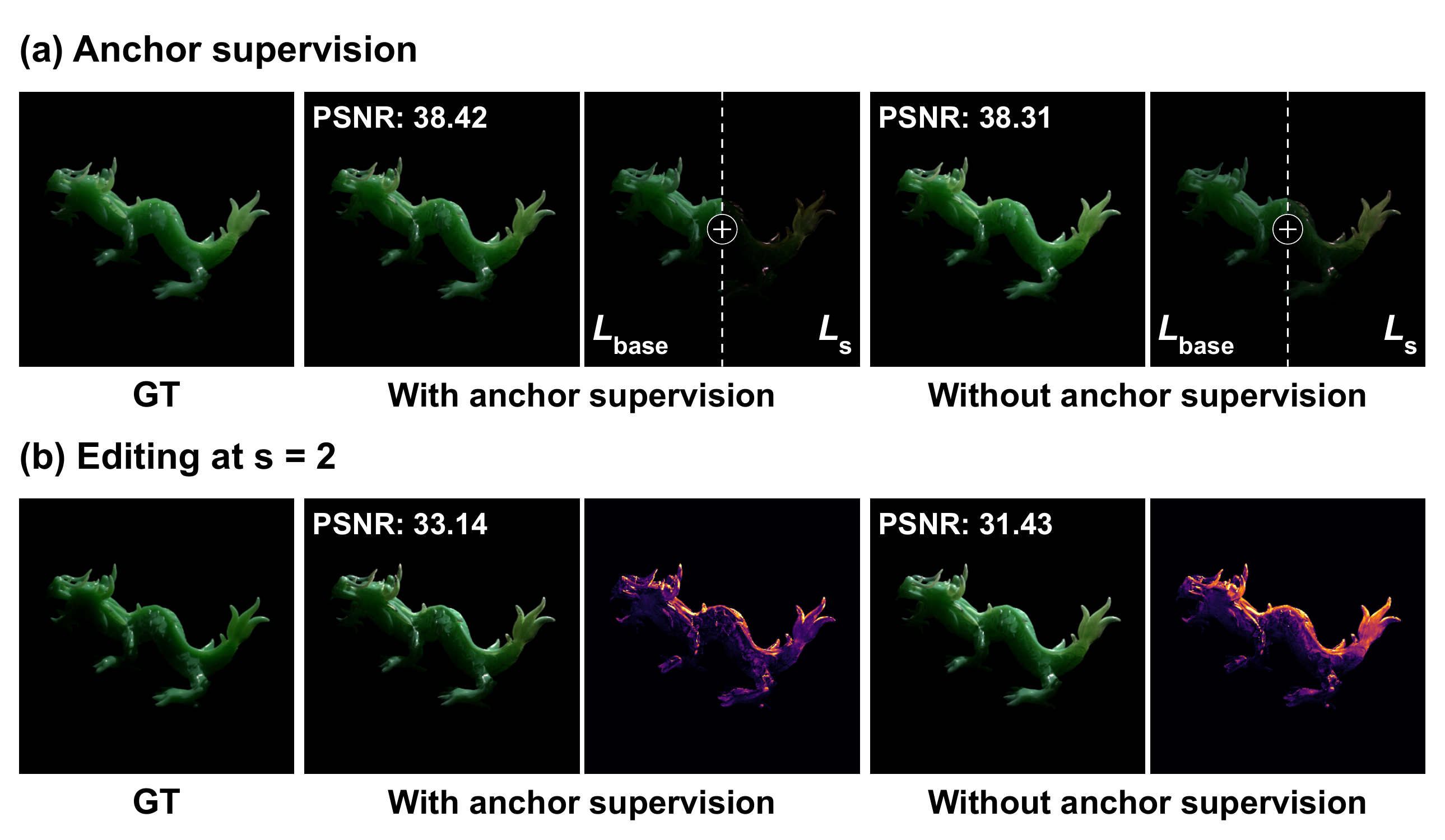}
    \caption{The comparison of results with/without anchor supervision. (a) Reconstructions into $L_s$ and $L_{\mathrm{base}}$. (b) Edited results at $s=2$. Anchor supervision improves component separation and agreement with the edited ground truth. }
    \label{fig:component_supervision_editing}
\end{figure}

%% file: sec_5_conclusion.tex
\section{Conclusion}
\label{sec:conclusion}
In this paper, we present NNS-GS, a framework for relighting translucent objects by learning non-local light transport. Our method extends local Gaussian evaluation to account for illumination across the object surface, producing more faithful scattering appearance under new lighting. Within this framework, geometry-guided reflection provides improved specular and shadow detail. Anchor-based progressive training keeps these contributions separate, supporting feasible scattering controls. Experiments on synthetic and real-world scenes demonstrate improved reconstruction and relighting quality over the compared methods, with better-preserved appearance details and real-time rendering.

A promising direction for future work is to extend NNS-GS to nearly transparent materials, where light passes through with little scattering. In these cases, surface illumination alone provides insufficient guidance for predicting the transmitted appearance. Accounting more directly for the direction of light traveling through the object could improve reconstruction and allow reliable editing over a wider range of transparency.

\section*{LLM Usage Statement}
We used ChatGPT and Codex solely to polish the manuscript text, refine the layout and visual presentation of selected figures, and modify the project code. All changes were reviewed by the authors, who take full responsibility for the final content.

%% file: supplementary_sec_implementation.tex
\section{Implementation Details}
\label{sec:supp_implementation}

\begin{figure}[!ht]
\centering
\includegraphics[width=\linewidth]{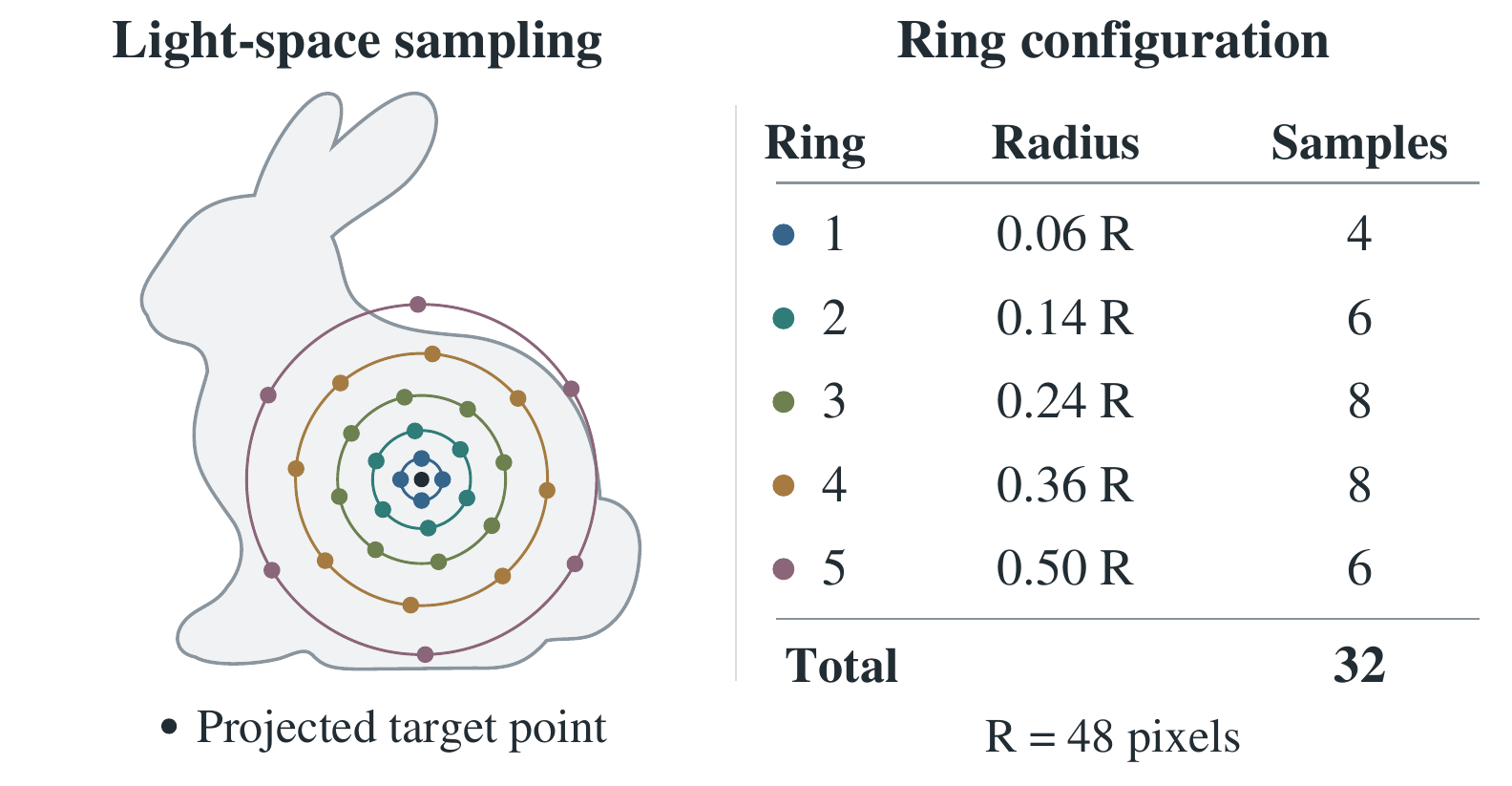}
\caption{Ring-based sampling in light space. Colored points show the 32 samples around the projected target point. Colors correspond to the five ring configurations on the right.}
\label{fig:supp_rings}
\end{figure}

\paragraph{Ring-based sampling.}
Section~3.3 of the main paper uses ring-based sampling in light space to aggregate non-local illumination cues. The layout below specifies how nearby detail and wider surface coverage are balanced under a fixed sampling budget. Since scattering contributions generally decrease with travel distance, $K=32$ samples are distributed over five concentric rings with progressively wider spacing. This provides finer coverage near the projected target and coarser coverage farther away. Fig.~\ref{fig:supp_rings} gives the empirical radii and sample counts used to achieve this balance. Samples are uniformly spaced along each ring, with adjacent rings angularly offset to cover different directions rather than repeatedly sampling the same radial lines.

\paragraph{Interpretation of the scattering scale.}
The scale $s$ in Sec.~3.3 of the main paper controls distance attenuation and diffuse--scattering blending. Its physical interpretation follows from the relation between distance and extinction. In a homogeneous medium with extinction coefficient $\sigma_t$, the transmittance over a path of length $\ell$ is $T=\exp(-\sigma_t\ell)$~\cite{pharr2023pbrt}. Scaling extinction by $s$ gives
\begin{equation}
 T_s=\exp[-(s\sigma_t)\ell]=\exp[-\sigma_t(s\ell)].
\end{equation}
Thus, scaling distance inside the exponential has the same effect on attenuation as scaling extinction, without changing the geometric path itself.

Our attenuation factor $\mathbf A$ in the main paper averages $\exp(-s d_j)$ over valid samples to modulate the aggregated scattering response, using the sample-to-target distance $d_j$ as a proxy for transport distance. Since $d_j$ is a Euclidean separation rather than a traced internal path, $s$ is interpreted relative to the reference setting $s=1$, not as a calibrated material coefficient. The same scale acts on the learned optical-thickness proxy through $\mathbf w=\exp(-s\boldsymbol\tau)$. For convenient control, this weight also blends scattering with the complementary diffuse contribution. These adjustments change attenuation and blending while keeping the learned transport response fixed.

\paragraph{Optimization parameters.}
The progressive optimization in Sec.~3.5 of the main paper is implemented with the following branch-specific learning rates and loss weights. All appearance networks use Adam~\cite{kingma2015adam} with an initial learning rate of $10^{-3}$ upon activation, except for shadow modulation, which uses $3\times10^{-4}$. The reconstruction loss combines $L_1$ and DSSIM with weights 0.8 and 0.2. The remaining weights are $\lambda_{\mathrm{TV}}=0.002$, $\lambda_{\mathrm{anchor}}=0.02$, and $\lambda_{\mathrm{spec}}=0.05$. We use $s=1$ except in the scattering-control experiments.

\paragraph{Network configuration.}
The scattering and reflection modules in Secs.~3.3--3.4 of the main paper use several networks with different roles. Their layer configurations are specified here according to the appearance variation handled by each branch. Scattering must represent transport across surface regions, whereas specular reflection must capture sharp directional changes. These branches therefore receive most of the network capacity, with smaller heads for the local frame, optical thickness, and shadow modulation. Tab.~\ref{tab:supp_networks} details this allocation. All appearance networks operate on 16-dimensional Gaussian latent codes and use ReLU in their hidden layers. The diffuse network is shared between light-space and view-space evaluation.

\begin{figure}[!tp]

\centering
\captionof{table}{Network configuration. Larger networks model scattering and specular appearance, while smaller heads estimate the auxiliary quantities. The listed widths refer to hidden layers only.}
\label{tab:supp_networks}
\small
\setlength{\tabcolsep}{4pt}
\begin{tabular}{ll}
\toprule
Network & Hidden widths\\
\midrule
Scattering & $256,256,256,256$\\
Optical thickness & $128,64$\\
Local-frame decoder & $64,32$\\
Diffuse & $128,128,64$\\
Specular & $256,256,256,256,128$\\
Shadow modulation & $32,32$\\
\bottomrule
\end{tabular}
\end{figure}

\begin{figure}[!htbp]
\centering
\captionsetup{hypcap=false}
\includegraphics[width=\linewidth]{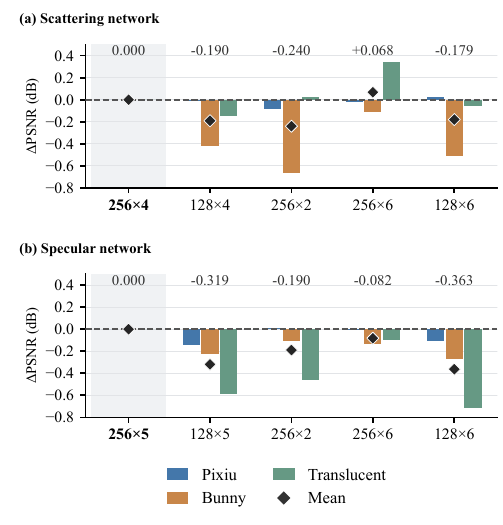}
\caption{Effect of scattering and specular network size. Narrower or shallower variants generally reduce quality, while deeper variants provide no consistent improvement across scenes. Each panel varies one network with the other fixed. $\Delta\mathrm{PSNR}$ (dB) is measured relative to the default configuration (zero baseline). Diamonds denote the three-scene mean.}
\label{fig:supp_network_capacity}
\end{figure}

The selected four-layer scattering and five-layer specular networks balance reconstruction quality against computational cost. To examine this choice, Fig.~\ref{fig:supp_network_capacity} compares alternative widths and depths on Pixiu, Bunny, and Translucent, selected from the three datasets for their complex details and translucent effects. Reducing width or depth generally lowers quality, while increasing capacity offers limited gains at a higher computational cost. These comparisons support the selected configuration. The specular label $256\times5$ is shorthand for hidden widths $256,256,256,256,128$.

\paragraph{Training schedule.}
\label{sec:supp_schedule}
The progressive strategy in Sec.~3.5 of the main paper introduces diffuse, scattering, and specular learning in sequence. The schedule comparison below examines how to allocate iterations among these stages without limiting final quality. We use 30k forward-splatting iterations and 70k deferred appearance iterations, similar to RNG~\cite{fan2025rng}. Within this fixed iteration budget, Tab.~\ref{tab:supp_schedule} compares allocations among diffuse, scattering, and specular learning to assess the quality--cost trade-off. Giving more iterations to earlier stages does not improve final quality in these comparisons. We select 10k/20k/40k, which achieves the best PSNR, SSIM, and LPIPS on all three scenes at a comparable training cost. The anchor is established after 30k appearance iterations, and the final 40k iterations refine the components with separate supervision.

\raggedbottom

\input{supplementary_sec_training_schedule.tex}

%% file: supplementary_sec_training_schedule.tex
\begin{figure}[!htbp]

\centering
\captionsetup{hypcap=false}\captionof{table}{Comparison of training-stage allocations under a fixed budget of 70k appearance-training iterations. The selected 10/20/40 schedule gives the best quality across the three scenes at comparable training cost. Allocations are in thousands of iterations. Time reports the measured appearance-training duration in minutes. Best quality metrics are bold.}
\label{tab:supp_schedule}
\begingroup
\footnotesize
\setlength{\tabcolsep}{2.5pt}
\renewcommand{\arraystretch}{1.12}
\begin{tabular*}{\linewidth}{@{\extracolsep{\fill}}lrrrr@{}}
\toprule
Schedule (k) & PSNR $\uparrow$ & SSIM $\uparrow$ & LPIPS $\downarrow$ & Time (min)\\
\midrule
\multicolumn{5}{l}{\textbf{Pixiu}}\\
\textbf{10/20/40} & \textbf{32.2203} & \textbf{0.9452} & \textbf{0.0709} & 59.43 \\
10/30/30 & 32.1313 & 0.9447 & 0.0716 & 56.62 \\
20/20/30 & 32.1118 & 0.9449 & 0.0719 & 55.03 \\
10/40/20 & 31.9836 & 0.9441 & 0.0727 & 51.47 \\
\midrule
\multicolumn{5}{l}{\textbf{Bunny}}\\
\textbf{10/20/40} & \textbf{41.8635} & \textbf{0.9922} & \textbf{0.0105} & 27.42 \\
10/30/30 & 41.7667 & 0.9920 & 0.0110 & 28.30 \\
20/20/30 & 41.5591 & 0.9918 & 0.0111 & 25.96 \\
10/40/20 & 41.3147 & 0.9915 & 0.0120 & 25.55 \\
\midrule
\multicolumn{5}{l}{\textbf{Translucent}}\\
\textbf{10/20/40} & \textbf{32.4912} & \textbf{0.9793} & \textbf{0.0229} & 62.01 \\
10/30/30 & 31.9516 & 0.9776 & 0.0239 & 58.12 \\
20/20/30 & 32.4350 & 0.9790 & 0.0234 & 55.97 \\
10/40/20 & 31.3533 & 0.9751 & 0.0265 & 57.05 \\
\bottomrule
\end{tabular*}
\endgroup
\end{figure}

%% file: supplementary_sec_additional_results.tex
\section{Additional Validation}
\label{sec:supp_validation}
\subsection{Quantitative Comparisons}
\label{sec:additional_results}

\begin{figure*}[!tp]

\centering
\captionof{table}{SSS-GS comparison under unseen light positions. Means give equal weight to each scene within the synthetic or real subset. Red and orange cells indicate the best and second-best results, respectively.}
\label{tab:supp_unseen_lights}
\begingroup
\footnotesize
\color{black}
\setlength{\tabcolsep}{1pt}
\renewcommand{\arraystretch}{1.15}
\begin{tabular*}{\textwidth}{@{\extracolsep{\fill}}l*{15}{r}@{}}
\toprule
Scene & \multicolumn{3}{c}{GS\textsuperscript{3}} & \multicolumn{3}{c}{SSD-GS} & \multicolumn{3}{c}{RNG} & \multicolumn{3}{c}{SSS-GS} & \multicolumn{3}{c}{Ours} \\
\cmidrule(lr){2-4}\cmidrule(lr){5-7}\cmidrule(lr){8-10}\cmidrule(lr){11-13}\cmidrule(lr){14-16}
 & PSNR $\uparrow$ & SSIM $\uparrow$ & LPIPS $\downarrow$ & PSNR $\uparrow$ & SSIM $\uparrow$ & LPIPS $\downarrow$ & PSNR $\uparrow$ & SSIM $\uparrow$ & LPIPS $\downarrow$ & PSNR $\uparrow$ & SSIM $\uparrow$ & LPIPS $\downarrow$ & PSNR $\uparrow$ & SSIM $\uparrow$ & LPIPS $\downarrow$ \\
\midrule

Bunny & 33.8300 & 0.9792 & 0.0264 & \cellcolor{orange!25}37.4300 & \cellcolor{orange!25}0.9882 & 0.0161 & 37.0700 & 0.9880 & \cellcolor{orange!25}0.0157 & 35.3100 & 0.9788 & 0.0258 & \cellcolor{red!20}40.8728 & \cellcolor{red!20}0.9919 & \cellcolor{red!20}0.0111 \\
Candle & 38.0700 & 0.9876 & 0.0167 & 40.4300 & 0.9917 & 0.0099 & \cellcolor{orange!25}42.4500 & \cellcolor{red!20}0.9953 & \cellcolor{red!20}0.0063 & 33.4400 & 0.9749 & 0.0288 & \cellcolor{red!20}44.0200 & \cellcolor{orange!25}0.9952 & \cellcolor{orange!25}0.0069 \\
Dragon & 35.2100 & 0.9739 & 0.0277 & \cellcolor{orange!25}36.8600 & \cellcolor{red!20}0.9802 & \cellcolor{red!20}0.0232 & 34.8200 & 0.9714 & 0.0336 & 34.8800 & 0.9628 & 0.0416 & \cellcolor{red!20}37.0900 & \cellcolor{orange!25}0.9791 & \cellcolor{orange!25}0.0251 \\
Soap & 37.4700 & 0.9771 & 0.0341 & 37.6900 & 0.9820 & 0.0241 & \cellcolor{orange!25}39.6200 & \cellcolor{orange!25}0.9845 & \cellcolor{red!20}0.0197 & 36.3800 & 0.9691 & 0.0311 & \cellcolor{red!20}40.8600 & \cellcolor{red!20}0.9859 & \cellcolor{red!20}0.0197 \\
Statue & 35.3100 & 0.9812 & 0.0213 & 34.2600 & 0.9781 & 0.0246 & \cellcolor{orange!25}37.0600 & \cellcolor{orange!25}0.9874 & \cellcolor{orange!25}0.0158 & 34.8800 & 0.9733 & 0.0306 & \cellcolor{red!20}39.0700 & \cellcolor{red!20}0.9887 & \cellcolor{red!20}0.0157 \\
\midrule
Mean (syn.) & 35.9780 & 0.9798 & 0.0252 & 37.3340 & 0.9840 & 0.0196 & \cellcolor{orange!25}38.2040 & \cellcolor{orange!25}0.9853 & \cellcolor{orange!25}0.0182 & 34.9780 & 0.9718 & 0.0316 & \cellcolor{red!20}40.3826 & \cellcolor{red!20}0.9882 & \cellcolor{red!20}0.0157 \\
\midrule
Clear Crystal & 34.7200 & 0.9792 & 0.0266 & 35.2700 & \cellcolor{orange!25}0.9798 & \cellcolor{orange!25}0.0226 & \cellcolor{orange!25}35.4600 & 0.9794 & 0.0236 & 34.4400 & 0.9763 & 0.0243 & \cellcolor{red!20}35.9600 & \cellcolor{red!20}0.9825 & \cellcolor{red!20}0.0177 \\
Green Candle & 30.3500 & 0.9573 & 0.0726 & 30.0600 & 0.9565 & 0.0788 & \cellcolor{orange!25}31.4800 & \cellcolor{orange!25}0.9629 & 0.0672 & 29.7300 & 0.9483 & \cellcolor{red!20}0.0634 & \cellcolor{red!20}32.6000 & \cellcolor{red!20}0.9646 & \cellcolor{orange!25}0.0649 \\
Joyful Yell & 30.8100 & 0.9616 & 0.0505 & 29.5200 & 0.9565 & 0.0559 & \cellcolor{orange!25}32.2000 & \cellcolor{orange!25}0.9677 & \cellcolor{orange!25}0.0403 & 31.0500 & 0.9589 & 0.0462 & \cellcolor{red!20}32.9700 & \cellcolor{red!20}0.9709 & \cellcolor{red!20}0.0340 \\
\midrule
Mean (real) & 31.9600 & 0.9660 & 0.0499 & 31.6167 & 0.9643 & 0.0524 & \cellcolor{orange!25}33.0467 & \cellcolor{orange!25}0.9700 & \cellcolor{orange!25}0.0437 & 31.7400 & 0.9612 & 0.0446 & \cellcolor{red!20}33.8433 & \cellcolor{red!20}0.9727 & \cellcolor{red!20}0.0389 \\
\bottomrule
\end{tabular*}
\endgroup
\end{figure*}

\input{supplementary_sec_nrhints_table.tex}

\input{supplementary_sec_gs3_table.tex}

\begin{figure*}[!tp]
\centering
\includegraphics[width=\textwidth,height=0.91\textheight,keepaspectratio]{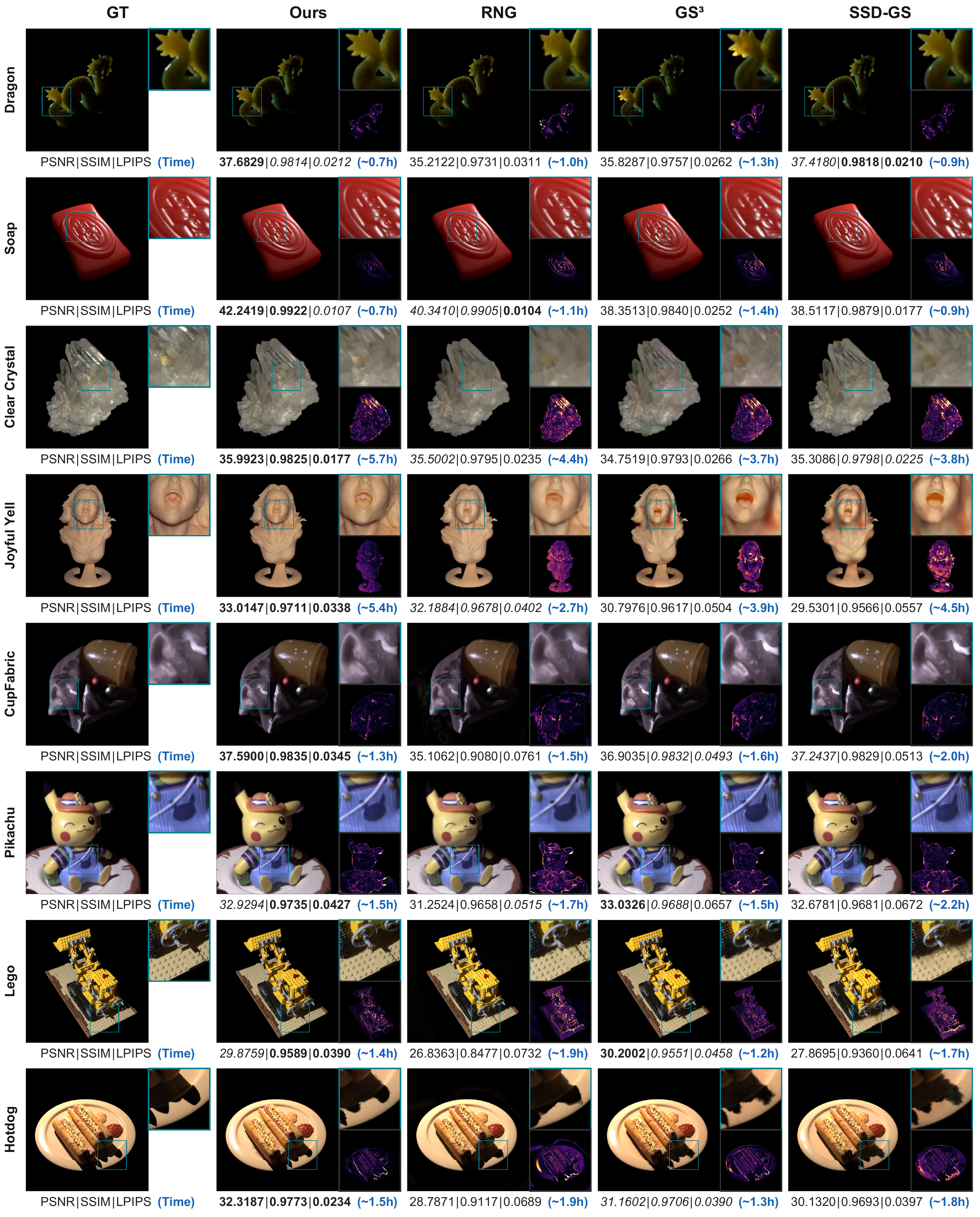}
\caption{Additional visual comparisons on further scenes from the evaluated datasets. Values report scene-average PSNR, SSIM, and LPIPS, followed by approximate training time in hours. Best and second-best metrics are bold and italic, respectively. For the SSS-GS scenes, the metrics are averages over the complete evaluation set reported in Tab.~1 of the main paper.}
\label{fig:supp_additional_comparison}
\end{figure*}

\paragraph{Additional evaluation.}
These scene-wise results extend Sec.~4.2 of the main paper to unseen lighting and broader material types. On SSS-GS~\cite{dihlmann2024sssgs} images with unseen light positions (Tab.~\ref{tab:supp_unseen_lights}), NNS-GS improves mean PSNR over RNG by 2.18~dB on synthetic scenes and 0.80~dB on real scenes, demonstrating improved relighting. On real NRHints~\cite{zeng2023nrhints} and synthetic GS\textsuperscript{3}~\cite{bi2024gs3}, mean PSNR improves by 0.75~dB over SSD-GS and 0.42~dB over GS\textsuperscript{3}, respectively (Tabs.~\ref{tab:nrhints_test} and~\ref{tab:gs3_test}). Together with the best mean SSIM and LPIPS on these datasets, these gains support applicability to a wider range of scenes, including those with fur and fine specular detail. However, metallic scenes remain challenging due to dominant local reflection and sharp anisotropic specular responses (Section~\ref{sec:limitations}). Fig.~\ref{fig:supp_additional_comparison} provides further visual comparisons on additional real and synthetic scenes from these datasets. Note that training takes longer on the large real-world scenes Clear Crystal and Joyful Yell because an anchor cache is constructed for every training image. Both cache construction and subsequent cache-reading I/O incur overhead that grows with training-set size and is included in the reported training times.

\begin{figure*}[!tp]

\centering
\captionof{table}{Comparison under environment lighting. NNS-GS and SSD-GS are trained under varying calibrated point lights, while GI-GS and GS-IR are trained under a fixed environment map. Red and orange cells indicate the best and second-best results, respectively.}
\label{tab:supp_env_comparison}
\begingroup
\footnotesize
\color{black}
\setlength{\tabcolsep}{1pt}
\renewcommand{\arraystretch}{1.0}
\begin{tabular*}{\textwidth}{@{\extracolsep{\fill}}l*{12}{r}@{}}
\toprule
Scene & \multicolumn{3}{c}{Ours} & \multicolumn{3}{c}{SSD-GS} & \multicolumn{3}{c}{GI-GS} & \multicolumn{3}{c}{GS-IR} \\
\cmidrule(lr){2-4}\cmidrule(lr){5-7}\cmidrule(lr){8-10}\cmidrule(lr){11-13}
 & PSNR $\uparrow$ & SSIM $\uparrow$ & LPIPS $\downarrow$ & PSNR $\uparrow$ & SSIM $\uparrow$ & LPIPS $\downarrow$ & PSNR $\uparrow$ & SSIM $\uparrow$ & LPIPS $\downarrow$ & PSNR $\uparrow$ & SSIM $\uparrow$ & LPIPS $\downarrow$ \\
\midrule
HairBlondePink & \cellcolor{red!20}33.6228 & \cellcolor{red!20}0.9428 & \cellcolor{red!20}0.0474 & \cellcolor{orange!25}32.6030 & 0.9260 & 0.0889 & 27.2115 & 0.9365 & 0.0599 & 27.0150 & \cellcolor{orange!25}0.9381 & \cellcolor{orange!25}0.0572 \\
WhiteFur & \cellcolor{red!20}31.2606 & \cellcolor{red!20}0.9284 & \cellcolor{red!20}0.0703 & \cellcolor{orange!25}29.4781 & 0.9012 & 0.1179 & 22.9559 & 0.9108 & 0.0845 & 23.6880 & \cellcolor{orange!25}0.9142 & \cellcolor{orange!25}0.0806 \\
Flowers & \cellcolor{red!20}32.6517 & \cellcolor{red!20}0.9492 & \cellcolor{red!20}0.0455 & \cellcolor{orange!25}30.5605 & 0.8949 & 0.1245 & 25.7829 & 0.9319 & 0.0546 & 26.9709 & \cellcolor{orange!25}0.9395 & \cellcolor{orange!25}0.0494 \\
Lego & \cellcolor{red!20}29.8414 & \cellcolor{red!20}0.9469 & \cellcolor{red!20}0.0404 & \cellcolor{orange!25}29.0350 & 0.9325 & 0.0866 & 24.0214 & 0.9306 & 0.0704 & 24.8031 & \cellcolor{orange!25}0.9329 & \cellcolor{orange!25}0.0680 \\
\midrule
Mean & \cellcolor{red!20}31.8441 & \cellcolor{red!20}0.9418 & \cellcolor{red!20}0.0509 & \cellcolor{orange!25}30.4192 & 0.9137 & 0.1045 & 24.9929 & 0.9275 & 0.0674 & 25.6193 & \cellcolor{orange!25}0.9312 & \cellcolor{orange!25}0.0638 \\
\bottomrule
\end{tabular*}
\endgroup
\end{figure*}

\begin{figure*}[!tp]
\centering
\includegraphics[width=\linewidth]{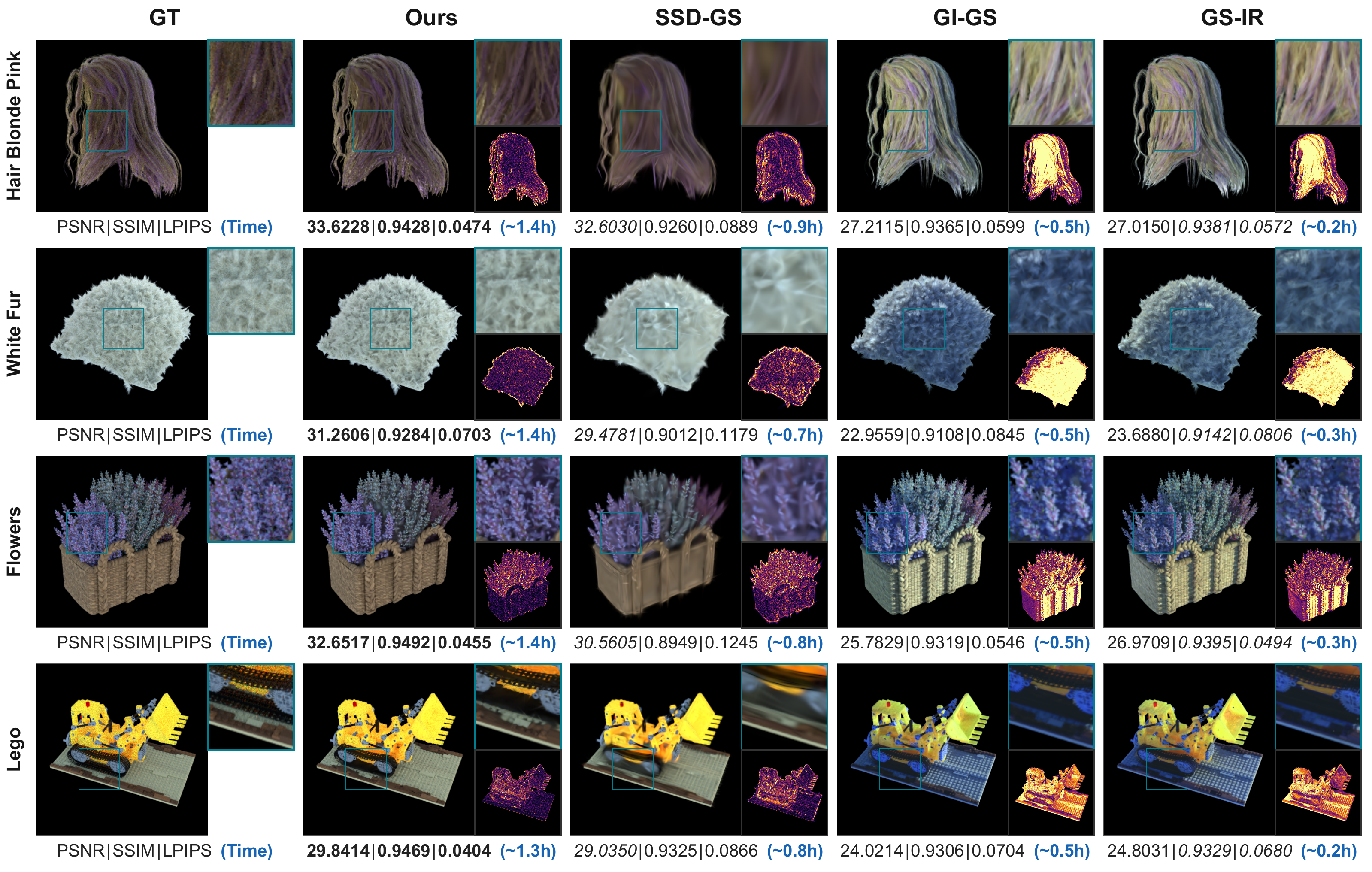}
\caption{Environment-relighting comparisons. Values report scene-average PSNR, SSIM, and LPIPS from Tab.~\ref{tab:supp_env_comparison}, followed by approximate training time in hours. Best and second-best metrics are bold and italic, respectively.}
\label{fig:supp_env_comparison}
\end{figure*}

\begingroup
\clubpenalties 4 10000 10000 10000 0
We further compare with GI-GS~\cite{chen2025gigs}, GS-IR~\cite{liang2024gsir}, and SSD-GS~\cite{zheng2026ssdgs} to evaluate relighting beyond point-light conditions. We approximate environment illumination through HDR-weighted integration of learned point-light responses without retraining, similar to RNG~\cite{fan2025rng}. All methods use the same scene assets, training camera poses, and evaluation protocol. NNS-GS and SSD-GS use varying point lights for training, while GI-GS and GS-IR use a fixed environment map. NNS-GS achieves the best scores across all three metrics on all four scenes (Tab.~\ref{tab:supp_env_comparison}). Fig.~\ref{fig:supp_env_comparison} shows closer agreement with the references across furry, floral, and structured objects. These improvements benefit from point-light supervision and the corresponding light-space cues, which help disentangle illumination from material properties, while non-local scattering captures contributions from other surface regions.
\par\endgroup

\begin{figure}[!htbp]

\centering
\captionof{table}{Additional ablations, grouped as baseline, scattering, reflection, and training. Full / K32 is the full model. Light center / K1 uses one light-space sample, while View-space sampling uses camera-view samples. The w/o distance input variant retains attenuation. Local dirs.\ w/o $h$ omits the half-vector, and World dirs.\ + $h$ omits the local-frame transform. Joint training removes progression and anchoring. Best values are bold.}
\label{tab:supp_sampling_half}
\begingroup
\footnotesize
\setlength{\tabcolsep}{3pt}
\renewcommand{\arraystretch}{1.08}
\begin{tabular*}{\columnwidth}{@{\extracolsep{\fill}}llrrr@{}}
\toprule
Scene & Configuration & PSNR $\uparrow$ & SSIM $\uparrow$ & LPIPS $\downarrow$ \\
\midrule

Translucent & Full / K32 & \textbf{32.4912} & \textbf{0.9793} & \textbf{0.0229} \\
\noalign{\vskip 2pt\hbox to \columnwidth{\cleaders\hbox{\rule{3pt}{0.35pt}\kern2pt}\hfill}\vskip 2pt}

 & Light center / K1 & 31.3279 & 0.9771 & 0.0240 \\

 & View-space sampling & 32.2190 & 0.9786 & 0.0246 \\

 & w/o distance input & 31.9240 & 0.9771 & 0.0251 \\
\noalign{\vskip 2pt\hbox to \columnwidth{\cleaders\hbox{\rule{3pt}{0.35pt}\kern2pt}\hfill}\vskip 2pt}

 & Local dirs. w/o $h$ & 31.6770 & 0.9766 & 0.0252 \\

 & World dirs. + $h$ & 31.7730 & 0.9776 & 0.0247 \\
\noalign{\vskip 2pt\hbox to \columnwidth{\cleaders\hbox{\rule{3pt}{0.35pt}\kern2pt}\hfill}\vskip 2pt}

 & Joint training & 31.9640 & 0.9708 & 0.0232 \\
\midrule

Pixiu & Full / K32 & \textbf{32.2203} & \textbf{0.9452} & 0.0709 \\
\noalign{\vskip 2pt\hbox to \columnwidth{\cleaders\hbox{\rule{3pt}{0.35pt}\kern2pt}\hfill}\vskip 2pt}

 & Light center / K1 & 31.8778 & 0.9439 & 0.0726 \\

 & View-space sampling & 32.0880 & 0.9447 & 0.0708 \\

 & w/o distance input & 32.1250 & 0.9446 & 0.0718 \\
\noalign{\vskip 2pt\hbox to \columnwidth{\cleaders\hbox{\rule{3pt}{0.35pt}\kern2pt}\hfill}\vskip 2pt}

 & Local dirs. w/o $h$ & 32.0210 & 0.9446 & 0.0716 \\

 & World dirs. + $h$ & 31.6900 & 0.9429 & 0.0745 \\
\noalign{\vskip 2pt\hbox to \columnwidth{\cleaders\hbox{\rule{3pt}{0.35pt}\kern2pt}\hfill}\vskip 2pt}

 & Joint training & 32.1120 & 0.9441 & \textbf{0.0699} \\
\midrule

Bunny & Full / K32 & 41.8635 & 0.9922 & \textbf{0.0105} \\
\noalign{\vskip 2pt\hbox to \columnwidth{\cleaders\hbox{\rule{3pt}{0.35pt}\kern2pt}\hfill}\vskip 2pt}

 & Light center / K1 & 40.3668 & 0.9901 & 0.0136 \\

 & View-space sampling & 39.4230 & 0.9883 & 0.0171 \\

 & w/o distance input & 40.0300 & 0.9892 & 0.0165 \\
\noalign{\vskip 2pt\hbox to \columnwidth{\cleaders\hbox{\rule{3pt}{0.35pt}\kern2pt}\hfill}\vskip 2pt}

 & Local dirs. w/o $h$ & 41.5890 & 0.9918 & 0.0112 \\

 & World dirs. + $h$ & 41.8469 & 0.9922 & 0.0106 \\
\noalign{\vskip 2pt\hbox to \columnwidth{\cleaders\hbox{\rule{3pt}{0.35pt}\kern2pt}\hfill}\vskip 2pt}

 & Joint training & \textbf{41.8830} & \textbf{0.9932} & 0.0108 \\
\bottomrule
\end{tabular*}
\endgroup
\end{figure}

\begin{figure}[!htbp]
\centering
\captionsetup{hypcap=false}
\includegraphics[width=\linewidth]{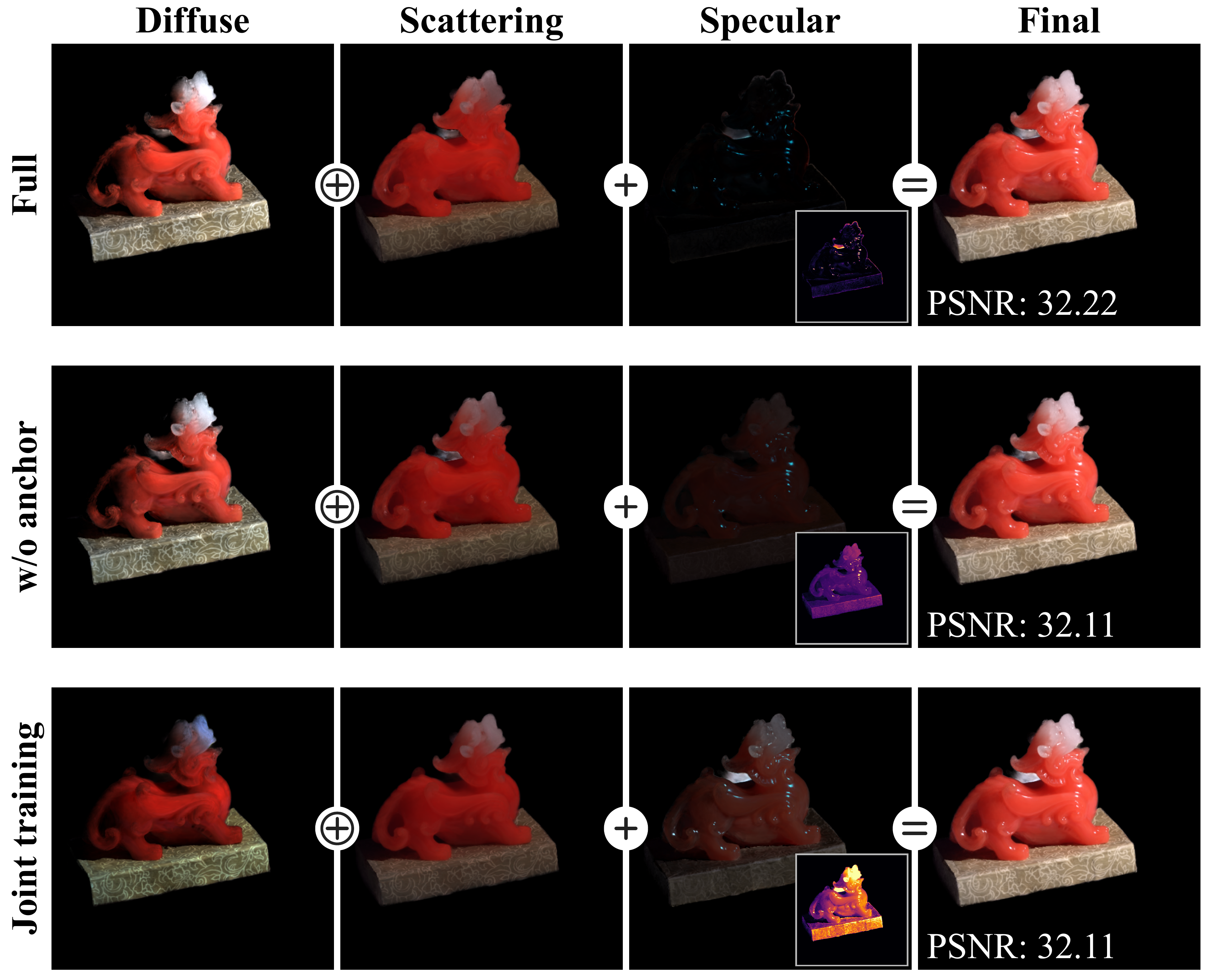}
\caption{Pixiu component separation. Circled plus denotes blending. Values are mean PSNR. Insets show intensity heatmaps on a shared scale.}
\label{fig:supp_pixiu_components}
\end{figure}

\input{supplementary_fig_validation_floats.tex}

\paragraph{Additional ablations.}
To validate the dual-space scattering evaluation in Sec.~3.3, we first retain only the target's projected point in light space and discard surrounding samples (Tab.~\ref{tab:supp_sampling_half}). The degradation across all three metrics supports aggregating illumination from multiple surrounding surface locations. We also collect cues in view space using the same ring-based sampling pattern with $K=32$. This can improve over local view-space prediction, but remains less effective than light-space sampling, since camera-visible samples can miss illuminated surfaces on the far side of the object, particularly under backlighting. \mbox{Finally}, removing distance inputs while retaining explicit attenuation lowers PSNR on all three scenes. This shows that spatial relationships, in addition to illumination cues, are important for learning scattering contributions.

The geometry-guided reflection module in Sec.~3.4 uses both a learned local frame and the half-vector $h$. We separate their effects by removing $h$ while retaining the frame, or using world-space directions with $h$. Both variants lower PSNR on all scenes. This supports their complementary roles in expressing directions relative to the surface and describing the light--view relationship for specular prediction.

To further examine the training strategy in Sec.~3.5, we compare joint training, progressive training without anchoring, and our full strategy. Although the three achieve similar final reconstruction quality, their component separation differs (Fig.~\ref{fig:supp_pixiu_components}). Conventional joint training allows the expressive specular branch to absorb diffuse and scattering contributions, producing incorrect separation. Progressive training~\cite{zheng2026ssdgs} reduces this mixing but does not fully resolve it in our comparison. Building on this progression, our strategy introduces a fixed anchor and supervises the components separately. This substantially reduces the absorption of base illumination into specular reflection, preserving the components that should respond to scattering control and making that control more consistent and meaningful.

%% file: supplementary_sec_nrhints_table.tex
\begin{figure*}[!tp]

\centering
\captionof{table}{Comparison on the real NRHints scenes. Means give equal weight to each scene. Red and orange cells indicate the best and second-best results, respectively.}
\label{tab:nrhints_test}
\begingroup
\footnotesize
\color{black}
\setlength{\tabcolsep}{1pt}
\renewcommand{\arraystretch}{1.15}
\begin{tabular*}{\textwidth}{@{\extracolsep{\fill}}l*{15}{r}@{}}
\toprule
Scene & \multicolumn{3}{c}{GS\textsuperscript{3}} & \multicolumn{3}{c}{SSD-GS} & \multicolumn{3}{c}{RNG} & \multicolumn{3}{c}{SSS-GS} & \multicolumn{3}{c}{Ours} \\
\cmidrule(lr){2-4}\cmidrule(lr){5-7}\cmidrule(lr){8-10}\cmidrule(lr){11-13}\cmidrule(lr){14-16}
 & PSNR $\uparrow$ & SSIM $\uparrow$ & LPIPS $\downarrow$ & PSNR $\uparrow$ & SSIM $\uparrow$ & LPIPS $\downarrow$ & PSNR $\uparrow$ & SSIM $\uparrow$ & LPIPS $\downarrow$ & PSNR $\uparrow$ & SSIM $\uparrow$ & LPIPS $\downarrow$ & PSNR $\uparrow$ & SSIM $\uparrow$ & LPIPS $\downarrow$ \\
\midrule
Cat & \cellcolor{orange!25}27.8042 & \cellcolor{red!20}0.9057 & \cellcolor{orange!25}0.1315 & 27.7860 & \cellcolor{orange!25}0.9039 & 0.1337 & 26.5336 & 0.8433 & 0.2009 & 16.1439 & 0.0627 & 0.3213 & \cellcolor{red!20}29.1121 & 0.9031 & \cellcolor{red!20}0.1138 \\
CatSmall & \cellcolor{orange!25}34.8576 & \cellcolor{red!20}0.9774 & 0.0643 & 34.6887 & 0.9759 & 0.0681 & 34.4486 & 0.9688 & \cellcolor{orange!25}0.0606 & 31.0645 & 0.9518 & 0.0729 & \cellcolor{red!20}35.8507 & \cellcolor{orange!25}0.9762 & \cellcolor{red!20}0.0496 \\
CupFabric & 36.9035 & \cellcolor{orange!25}0.9832 & \cellcolor{orange!25}0.0493 & \cellcolor{orange!25}37.2437 & 0.9829 & 0.0513 & 35.1062 & 0.9080 & 0.0761 & 31.8768 & 0.9621 & 0.0611 & \cellcolor{red!20}37.5900 & \cellcolor{red!20}0.9835 & \cellcolor{red!20}0.0345 \\
Fish & \cellcolor{orange!25}31.0797 & \cellcolor{orange!25}0.9244 & 0.0879 & \cellcolor{red!20}31.1446 & \cellcolor{red!20}0.9255 & \cellcolor{orange!25}0.0870 & 29.2087 & 0.8926 & 0.1285 & 24.1757 & 0.8232 & 0.1401 & 30.9468 & 0.9190 & \cellcolor{red!20}0.0849 \\
FurScene & 31.2540 & 0.9536 & 0.0678 & \cellcolor{orange!25}31.5472 & \cellcolor{orange!25}0.9562 & \cellcolor{orange!25}0.0672 & 27.6010 & 0.9136 & 0.1116 & 19.8663 & 0.7805 & 0.2002 & \cellcolor{red!20}32.6979 & \cellcolor{red!20}0.9585 & \cellcolor{red!20}0.0587 \\
Pikachu & \cellcolor{red!20}33.0326 & \cellcolor{orange!25}0.9688 & 0.0657 & 32.6781 & 0.9681 & 0.0672 & 31.2524 & 0.9658 & \cellcolor{orange!25}0.0515 & 27.8367 & 0.9357 & 0.0770 & \cellcolor{orange!25}32.9294 & \cellcolor{red!20}0.9735 & \cellcolor{red!20}0.0427 \\
Pixiu & 30.9928 & 0.9433 & 0.0807 & \cellcolor{orange!25}31.0164 & \cellcolor{orange!25}0.9448 & \cellcolor{orange!25}0.0804 & 28.4480 & 0.8788 & 0.1291 & 23.4342 & 0.8532 & 0.1247 & \cellcolor{red!20}32.2203 & \cellcolor{red!20}0.9452 & \cellcolor{red!20}0.0709 \\
\midrule
Mean & 32.2749 & 0.9509 & \cellcolor{orange!25}0.0782 & \cellcolor{orange!25}32.3007 & \cellcolor{orange!25}0.9511 & 0.0793 & 30.3712 & 0.9101 & 0.1083 & 24.9140 & 0.7670 & 0.1425 & \cellcolor{red!20}33.0496 & \cellcolor{red!20}0.9513 & \cellcolor{red!20}0.0650 \\
\bottomrule
\end{tabular*}
\endgroup
\end{figure*}

%% file: supplementary_sec_gs3_table.tex
\begin{figure*}[!tp]

\centering
\captionof{table}{Comparison on the synthetic GS\textsuperscript{3} scenes. Means give equal weight to each scene. Red and orange cells indicate the best and second-best results, respectively.}
\label{tab:gs3_test}
\begingroup
\footnotesize
\color{black}
\setlength{\tabcolsep}{1pt}
\renewcommand{\arraystretch}{1.0}
\begin{tabular*}{\textwidth}{@{\extracolsep{\fill}}l*{15}{r}@{}}
\toprule
Scene & \multicolumn{3}{c}{GS\textsuperscript{3}} & \multicolumn{3}{c}{SSD-GS} & \multicolumn{3}{c}{RNG} & \multicolumn{3}{c}{SSS-GS} & \multicolumn{3}{c}{Ours} \\
\cmidrule(lr){2-4}\cmidrule(lr){5-7}\cmidrule(lr){8-10}\cmidrule(lr){11-13}\cmidrule(lr){14-16}
 & PSNR $\uparrow$ & SSIM $\uparrow$ & LPIPS $\downarrow$ & PSNR $\uparrow$ & SSIM $\uparrow$ & LPIPS $\downarrow$ & PSNR $\uparrow$ & SSIM $\uparrow$ & LPIPS $\downarrow$ & PSNR $\uparrow$ & SSIM $\uparrow$ & LPIPS $\downarrow$ & PSNR $\uparrow$ & SSIM $\uparrow$ & LPIPS $\downarrow$ \\
\midrule
AnisoMetal & \cellcolor{red!20}28.2206 & \cellcolor{red!20}0.9602 & \cellcolor{orange!25}0.0372 & \cellcolor{orange!25}27.7980 & \cellcolor{orange!25}0.9570 & 0.0395 & 25.2223 & 0.9338 & 0.0529 & 21.7355 & 0.8939 & 0.0728 & 27.7111 & 0.9553 & \cellcolor{red!20}0.0369 \\
Drums & \cellcolor{red!20}33.2668 & \cellcolor{red!20}0.9765 & \cellcolor{red!20}0.0318 & 32.5258 & \cellcolor{orange!25}0.9724 & \cellcolor{orange!25}0.0388 & 30.5904 & 0.9697 & 0.0432 & 29.6458 & 0.9507 & 0.0568 & \cellcolor{orange!25}32.6456 & 0.9721 & 0.0464 \\
FurBall & \cellcolor{orange!25}33.5716 & \cellcolor{red!20}0.9684 & 0.0520 & 33.0947 & 0.9625 & 0.0595 & 32.2116 & 0.9481 & 0.0664 & 33.3570 & 0.9564 & \cellcolor{orange!25}0.0496 & \cellcolor{red!20}35.3251 & \cellcolor{orange!25}0.9666 & \cellcolor{red!20}0.0444 \\
Hotdog & \cellcolor{orange!25}31.1602 & \cellcolor{orange!25}0.9706 & \cellcolor{orange!25}0.0390 & 30.1320 & 0.9693 & 0.0397 & 28.7871 & 0.9117 & 0.0689 & 26.6642 & 0.9418 & 0.0585 & \cellcolor{red!20}32.3187 & \cellcolor{red!20}0.9773 & \cellcolor{red!20}0.0234 \\
Lego & \cellcolor{red!20}30.2002 & \cellcolor{orange!25}0.9551 & \cellcolor{orange!25}0.0458 & 27.8695 & 0.9360 & 0.0641 & 26.8363 & 0.8477 & 0.0732 & 24.5831 & 0.8714 & 0.1081 & \cellcolor{orange!25}29.8759 & \cellcolor{red!20}0.9589 & \cellcolor{red!20}0.0390 \\
Translucent & \cellcolor{orange!25}31.4490 & \cellcolor{orange!25}0.9774 & 0.0270 & 30.4825 & 0.9772 & \cellcolor{orange!25}0.0243 & 27.5820 & 0.9578 & 0.0426 & 28.2642 & 0.9593 & 0.0399 & \cellcolor{red!20}32.4912 & \cellcolor{red!20}0.9793 & \cellcolor{red!20}0.0229 \\
\midrule
Mean & \cellcolor{orange!25}31.3114 & \cellcolor{orange!25}0.9680 & \cellcolor{orange!25}0.0388 & 30.3171 & 0.9624 & 0.0443 & 28.5383 & 0.9281 & 0.0579 & 27.3750 & 0.9289 & 0.0643 & \cellcolor{red!20}31.7279 & \cellcolor{red!20}0.9682 & \cellcolor{red!20}0.0355 \\
\bottomrule
\end{tabular*}
\endgroup
\end{figure*}

%% file: supplementary_fig_validation_floats.tex
\begin{figure}[!t]
\centering

\includegraphics[width=\linewidth]{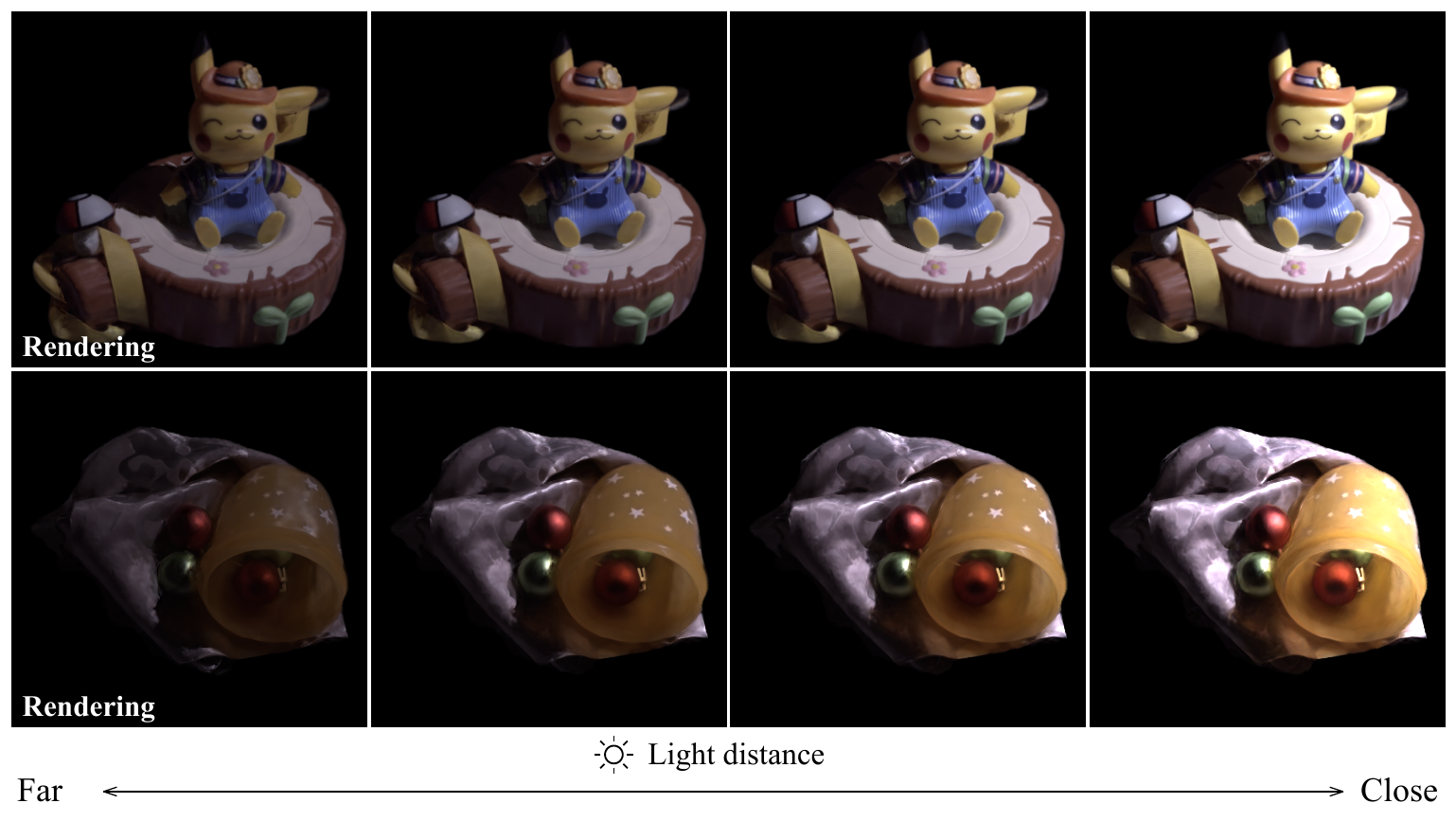}
\caption{Relighting as the light moves from far to close. The model captures changes in illumination and local specular appearance while the camera remains fixed.}
\label{fig:supp_light_distance}
\end{figure}

\begin{figure*}[!tp]
\centering
    \includegraphics[width=\linewidth]{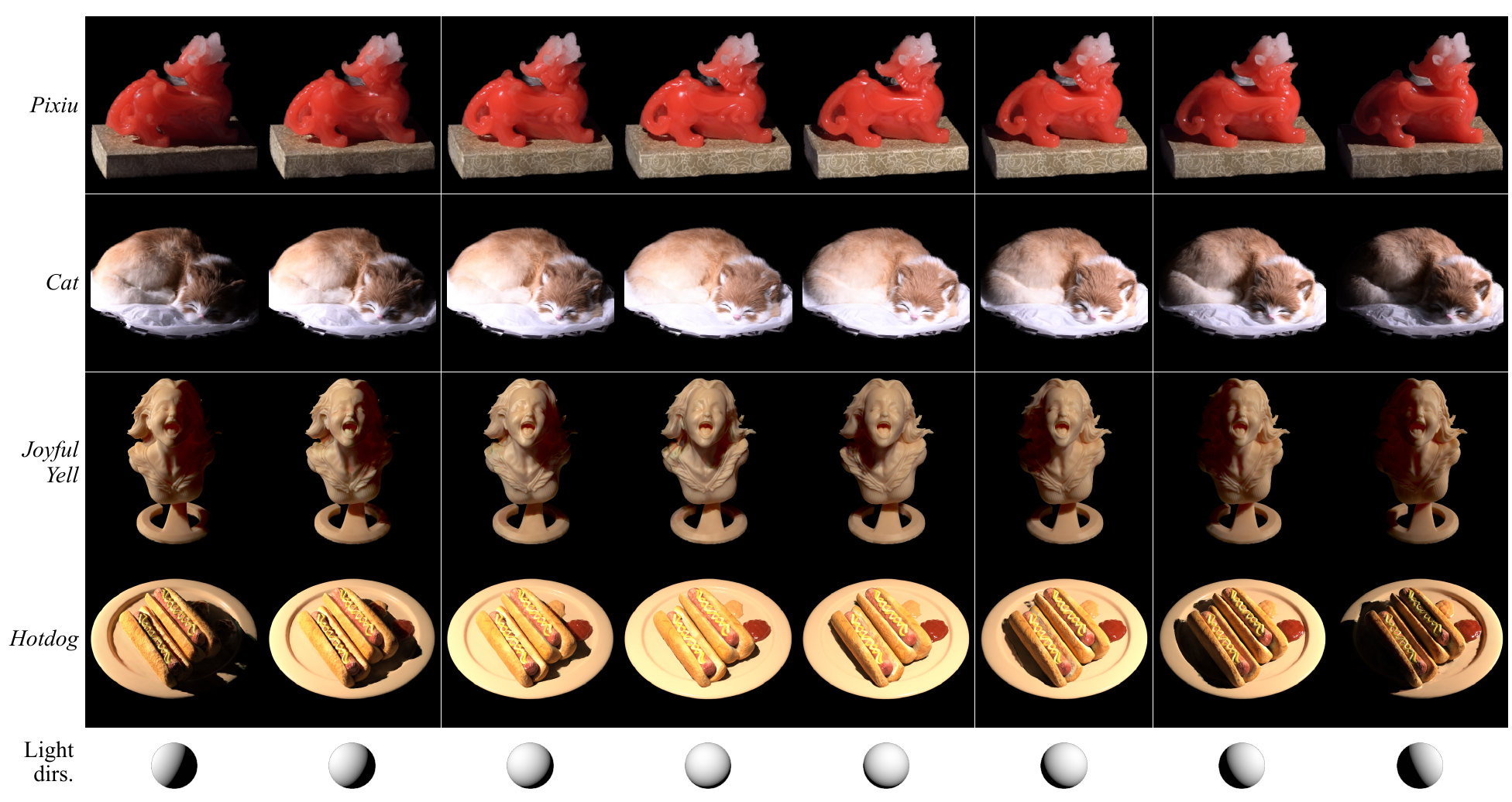}
    \caption{Relighting under moving point lights. Changes in illumination produce corresponding variations in shading, specular detail, and self-shadowing across the four scenes. The camera is fixed, and the spheres indicate the light directions.}
    \label{fig:supp_multilight}
\end{figure*}

\begin{figure*}[!tp]
\centering

\includegraphics[width=\linewidth,trim=-4.960bp 0bp 5.759bp 0bp]{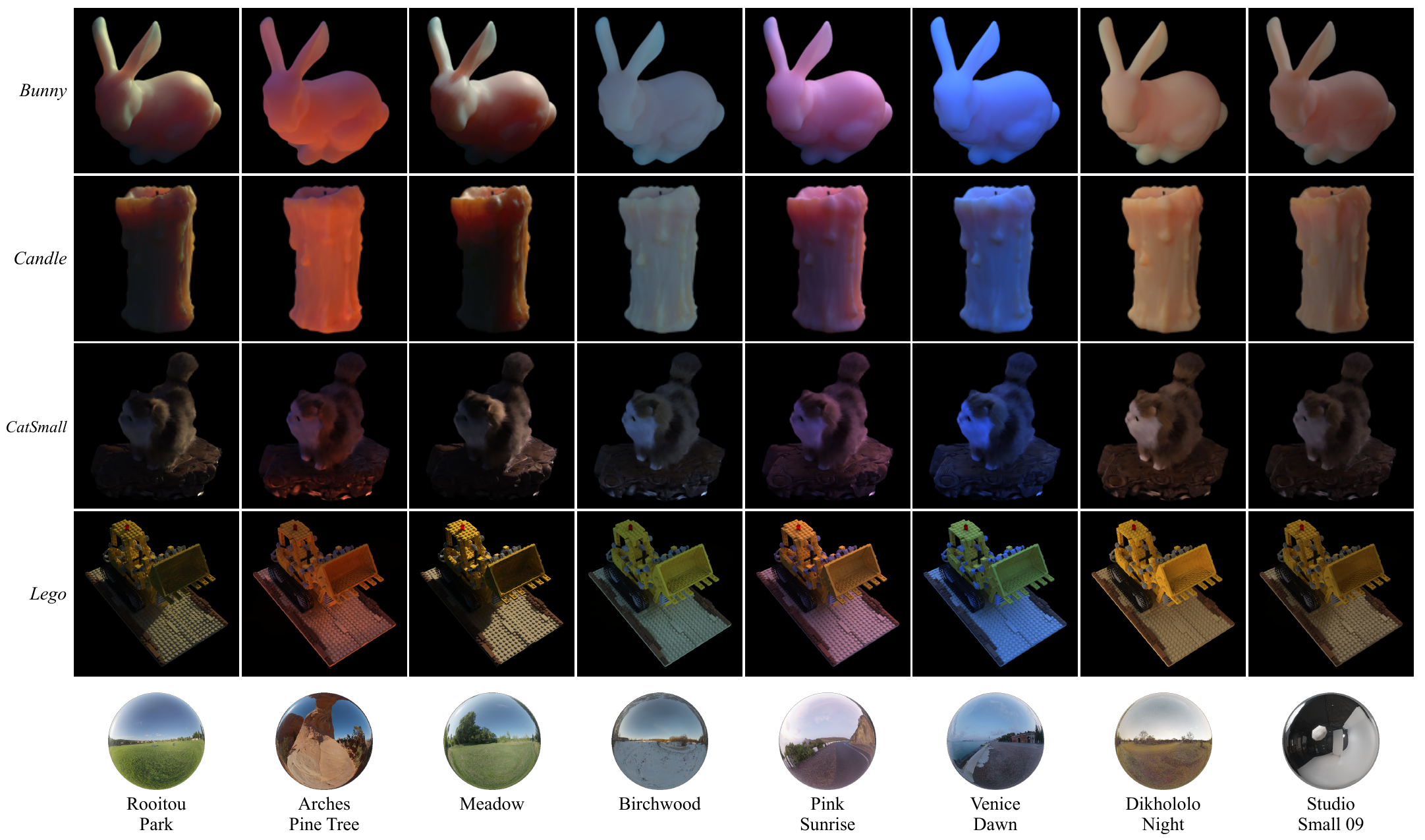}
\caption{Relighting under eight environment maps. The results demonstrate appearance responses to illumination with different colors and directional distributions across four types of objects. The spheres show the corresponding environments.}
\label{fig:supp_environment}
\end{figure*}

\begin{figure*}[!tp]
\centering

\includegraphics[width=\linewidth,trim=36.766bp 15bp 0bp 0bp]{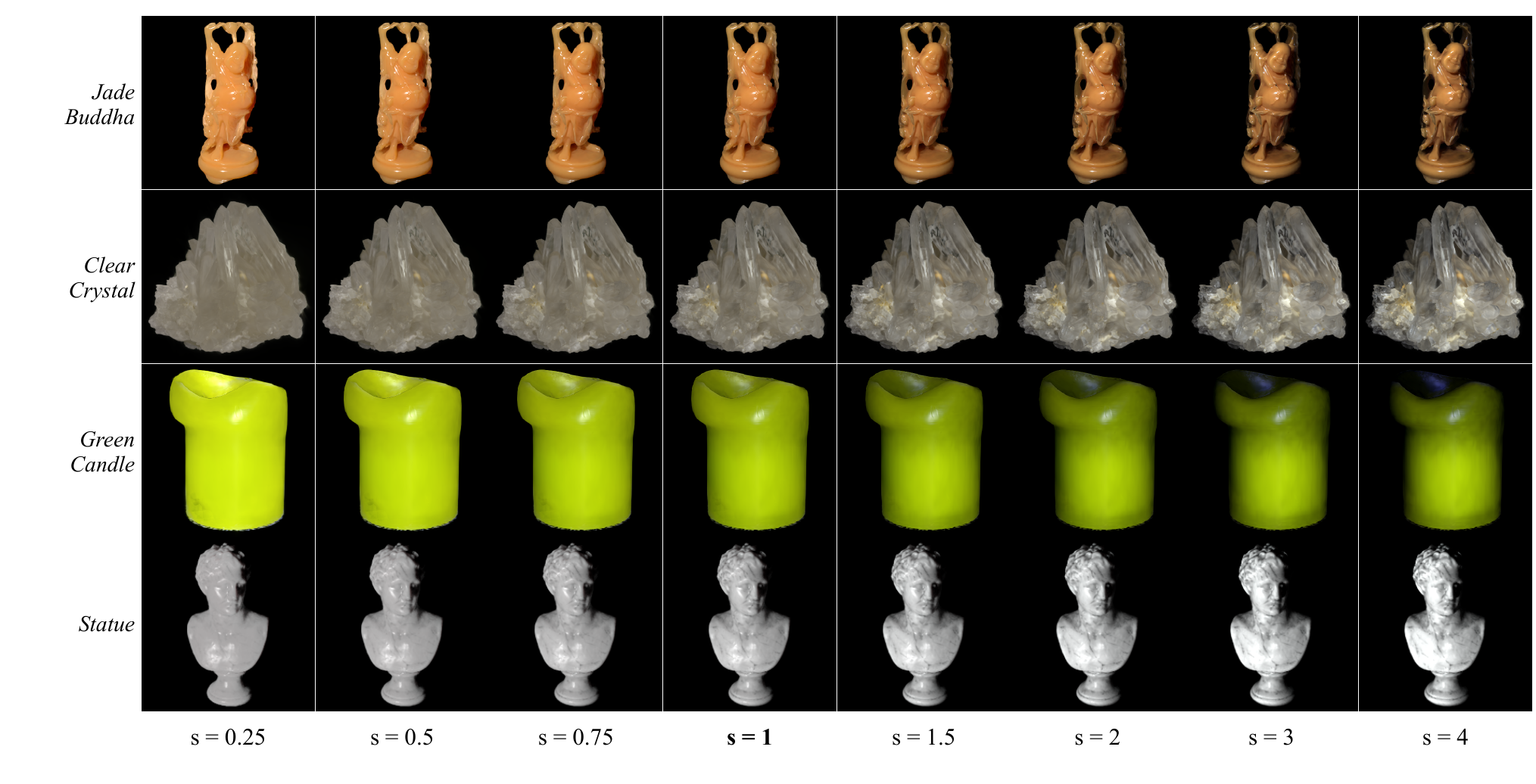}
    \caption{Scattering appearance control by varying $s$. Increasing $s$ attenuates the modeled scattering contribution and favors diffuse reflection, sharpening shading and shadow transitions. Camera and lighting are fixed, and $s=1$ is the default.}
    \label{fig:supp_scattering_edit}
\end{figure*}

%% file: supplementary_sec_relighting.tex
\subsection{Relighting}
\label{sec:supp_relighting}
\paragraph{Light distance.}
Fig.~\ref{fig:supp_light_distance} shows how appearance changes as a point light moves closer to the object while the camera remains fixed. The model captures increased \mbox{illumination} together with changes in the distribution of specular \mbox{reflections} across the surface. These results show a response to the spatial influence of light-source distance, rather than uniform brightness scaling alone.

\paragraph{Light direction.}
Fig.~\ref{fig:supp_multilight} shows relighting under different point-light directions with the camera fixed. Changes in light direction produce corresponding changes in shading, specular patterns, and self-shadowing across translucent, \mbox{reflective}, and fibrous surfaces. The results illustrate how our representation captures both smooth appearance transitions and sharp lighting details across these materials.

\paragraph{Environment lighting.}
Fig.~\ref{fig:supp_environment} shows a range of objects relit under multiple HDR environment maps, complementing the comparisons in Sec.~\ref{sec:additional_results}. The rendered colors and shading respond to changes in the color and directional distribution of illumination while retaining the distinct appearance of translucent, furry, and predominantly reflective objects. These results illustrate the extension of our point-light representation to varied environment lighting.

%% file: supplementary_sec_editing.tex
\subsection{Scattering Appearance Control}\label{sec:supp_editing}
The main paper compares scattering control with reference appearances at selected settings. The sequences here extend this comparison by showing how appearance varies across a wider range of $s$. With the camera and lighting fixed in Fig.~\ref{fig:supp_scattering_edit}, smaller values produce softer scattering appearance, while larger values reveal more distinct shading and shadow boundaries. This variation in spatial contrast goes beyond uniform brightness scaling and illustrates a transition toward diffuse reflection without retraining.

\medskip

%% file: supplementary_sec_limitations.tex
\begin{figure}[!t]
\centering
\captionsetup{hypcap=false}
\includegraphics[width=\linewidth]{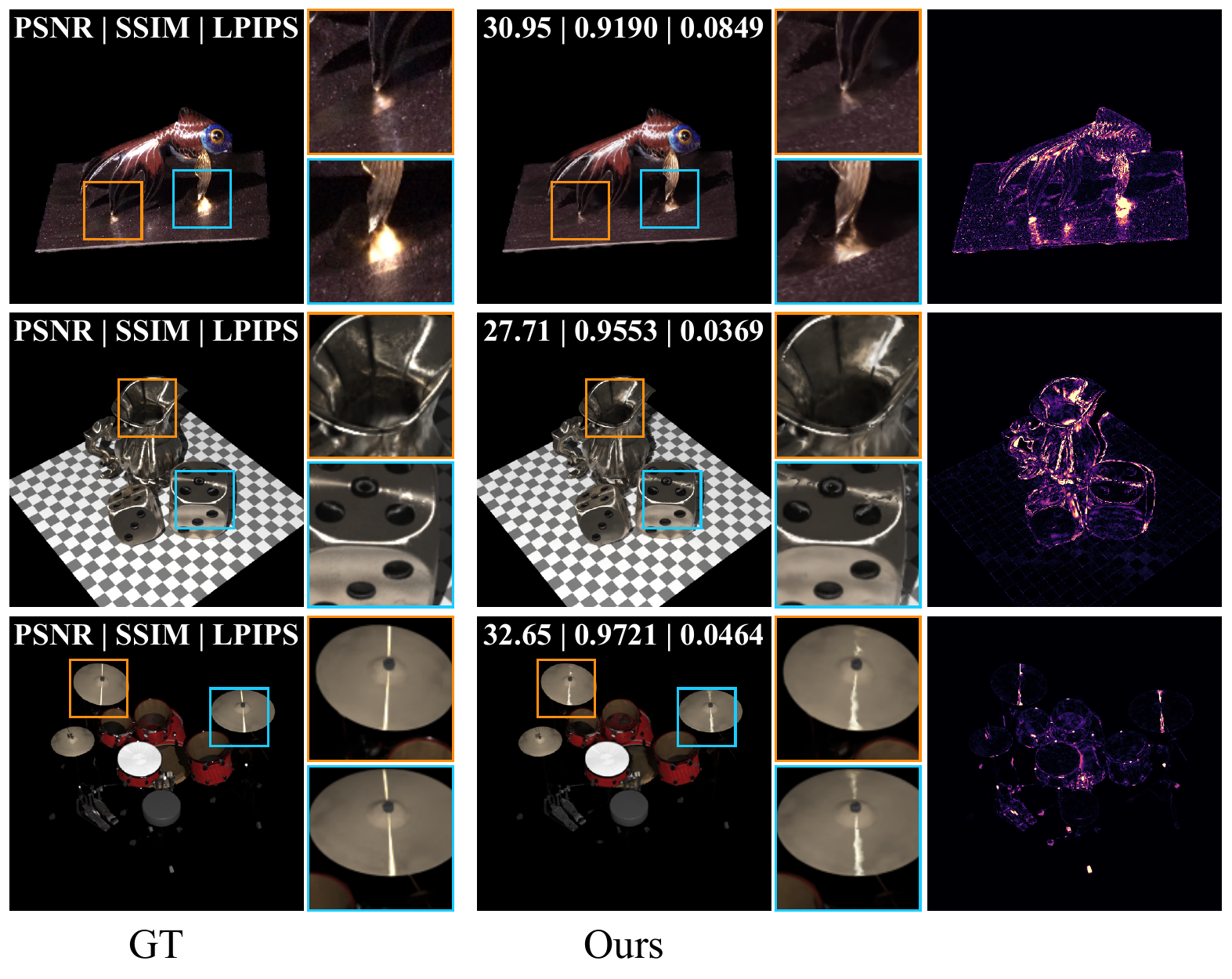}
\caption{Remaining errors in metallic specular detail. Enlarged regions and error maps highlight differences from the ground truth. Metrics are scene averages for our method.}
\label{fig:supp_failure_cases}
\end{figure}

\begin{samepage}
\section{Discussion and Future Work}
\label{sec:limitations}

Reproducing sharp specular detail on metallic surfaces remains challenging for our method. Fig.~\ref{fig:supp_failure_cases} shows errors relative to ground truth on Fish, AnisoMetal, and Drums (top to bottom). Non-local scattering offers limited benefit for metals because their appearance is dominated by reflection at the local surface point. Their strongly anisotropic specular responses change rapidly with light and view direction, remaining challenging despite our local-frame representation and dedicated supervision. We leave reflection representations that better capture directional changes and fine specular detail to future work.
\par
\end{samepage}

\begin{figure}[!t]
\centering
\captionsetup{hypcap=false}
\includegraphics[width=\linewidth]{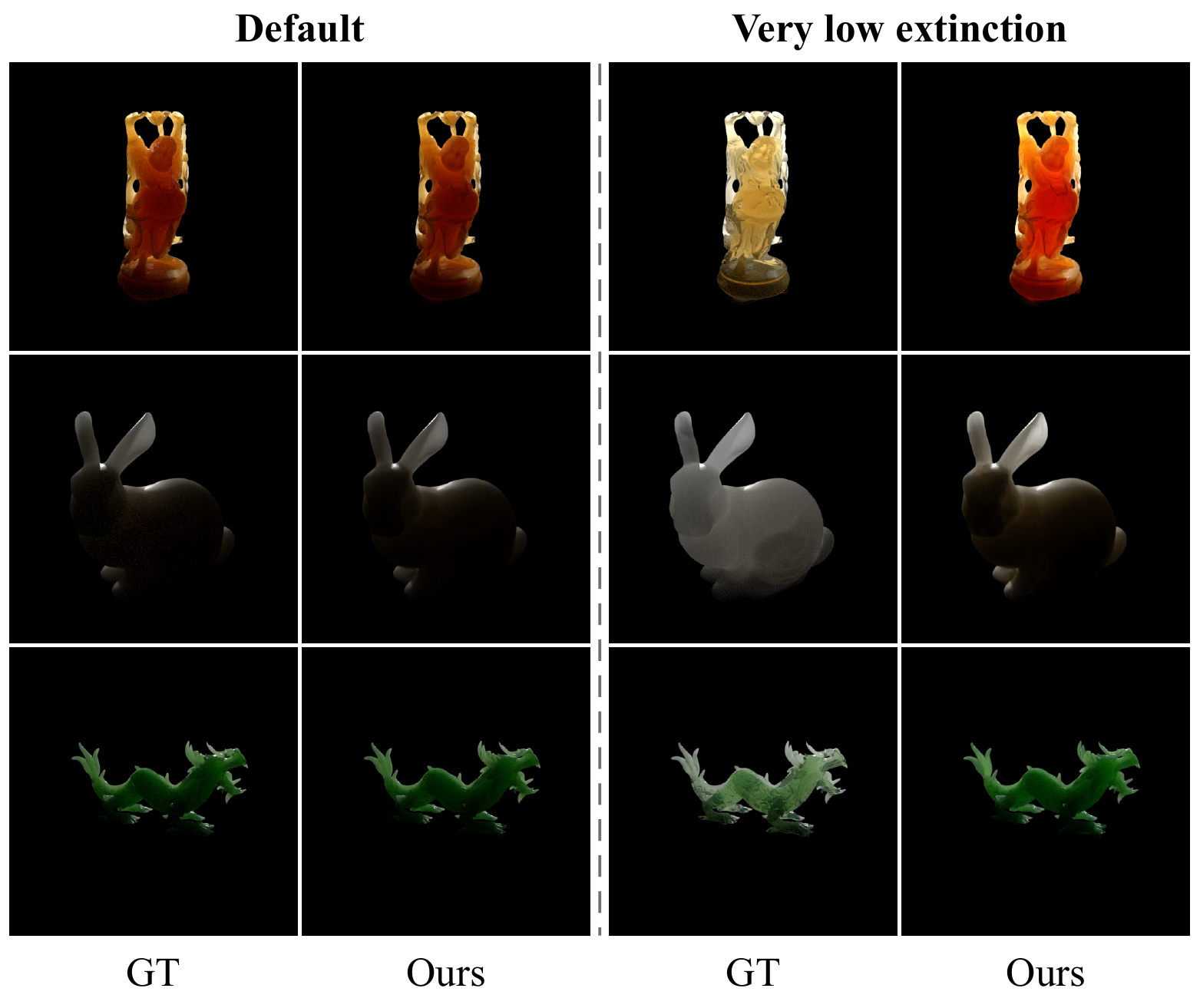}
\caption{Toward scattering control at very low extinction. The left columns show the default setting ($s=1$). In the right columns, all three examples use $s<0.2$. Differences in color and brightness from the more transmissive reference motivate further work on transport cues, sampling, and parameter calibration.}
\label{fig:supp_low_extinction}
\end{figure}

Extending scattering control to more transparent appearances is another promising direction for future work. When $s$ is reduced to a very low value, our results differ from the reference in brightness and color (Fig.~\ref{fig:supp_low_extinction}). Reducing $s$ adjusts attenuation but leaves the learned transport relationships unchanged. In a real material, however, lower extinction can change both the paths of light through the object and which surface regions contribute to its appearance. These changes may exceed the transport variations seen during training. With fewer scattering events, appearance depends more strongly on the incoming light direction, which our diffuse cues do not fully describe. Our fixed sampling pattern may also miss relevant contributions as these regions change. Moreover, the learned optical-thickness proxy is not a calibrated material parameter, so changing $s$ does not necessarily match a physical material change. Future work could address these gaps through richer directional cues, adaptive sampling, and a closer correspondence between the control parameter and physical material properties. These extensions could support a broader range of translucent materials and more precise control over their scattering appearance.